\documentclass{article}

\usepackage[final, main, nonatbib]{neurips_2026} % camera-ready

\usepackage[utf8]{inputenc} % allow utf-8 input
\usepackage[T1]{fontenc}    % use 8-bit T1 fonts
\usepackage{hyperref}       % hyperlinks
\usepackage{url}            % simple URL typesetting
\usepackage{booktabs}       % professional-quality tables
\usepackage{amsfonts}       % blackboard math symbols
\usepackage{nicefrac}       % compact symbols for 1/2, etc.
\usepackage{microtype}      % microtypography
\usepackage{xcolor}
\usepackage{subcaption}
\usepackage{lipsum}
\usepackage{amsmath}
\usepackage{multirow}
\usepackage[table]{xcolor}
\usepackage[normalem]{ulem}
\usepackage{adjustbox}
\usepackage{algorithm}
\usepackage{algpseudocode}
\usepackage{graphicx}
\usepackage{wrapfig}
\usepackage{float}
\usepackage{xcolor}

\definecolor{orange}{RGB}{242,170,107}
\definecolor{sky}{RGB}{107,218,242}
\definecolor{red}{RGB}{255,0,0}
\definecolor{blue}{RGB}{0,0,255}
\definecolor{rred}{RGB}{245, 152, 153}
\definecolor{oorange}{RGB}{253, 205, 154}
\definecolor{yyellow}{RGB}{248,244,140}

\title{DySurface: Consistent 4D Surface Reconstruction via Bridging Explicit Gaussians and Implicit Functions}

\author{
  Minje Kim$^{1}$ \quad
  Younghyun Noh$^{2}$ \quad
  Jaesoon Kim$^{1}$ \quad
  Tae-Kyun Kim$^{1}$ \\
  \\
  $^{1}$KAIST \quad
  $^{2}$KT  \\
  {\tt\small \{minjekim, kimtaekyun\}@kaist.ac.kr} \quad
  {\tt\small younghyun.noh@kt.com} \quad
  {\tt\small winnercup@g.skku.edu}
}

\begin{document}
\maketitle

\vspace{-0.1in}

\begin{abstract}
While novel view synthesis (NVS) for dynamic scenes has seen significant progress, reconstructing temporally consistent geometric surfaces remains a challenge. Neural Radiance Fields (NeRF) and 3D Gaussian Splatting (3DGS) offer powerful dynamic scene rendering capabilities; however, relying solely on photometric optimization often leads to geometric ambiguities. This results in discontinuous surfaces, severe artifacts, and broken surfaces over time. To address these limitations, we present DySurface, a novel framework that bridges the effectiveness of explicit Gaussians with the geometric fidelity of implicit Signed Distance Functions (SDFs) in dynamic scenes. Our approach tackles the structural discrepancy between the forward deformation of 3DGS ($canonical \rightarrow dynamic$) and the backward deformation required for volumetric SDF rendering ($dynamic \rightarrow canonical$). Specifically, we propose the VoxGS-DSDF branch that leverages deformed Gaussians to construct a dynamic sparse voxel grid, providing explicit geometric guidance to the implicit SDF field. This explicit anchoring effectively regularizes the volumetric rendering process, significantly improving surface reconstruction quality, with watertight boundaries and detailed representations. Quantitative and qualitative experiments demonstrate that DySurface significantly outperforms state-of-the-art baselines in geometric accuracy while maintaining competitive rendering performance. Project Page: \url{https://yunminjin2.github.io/projects/dysurface}.
\end{abstract}
% The obtained personalized super-resolution modules and 3D hand models are valuable for dynamic VR/AR systems and real-time applications. 
% Prior generative models have leveraged 3D reconstruction to enforce multi-view consistency on output images.
% While maintaining high-frequency details in 2D/3D is challenging, recent methods have achieved detailed 3D reconstruction from high-resolution multi-view videos. 

\vspace{-0.2in}
\begin{figure}[h]
\centering
\includegraphics[width=0.83\textwidth]{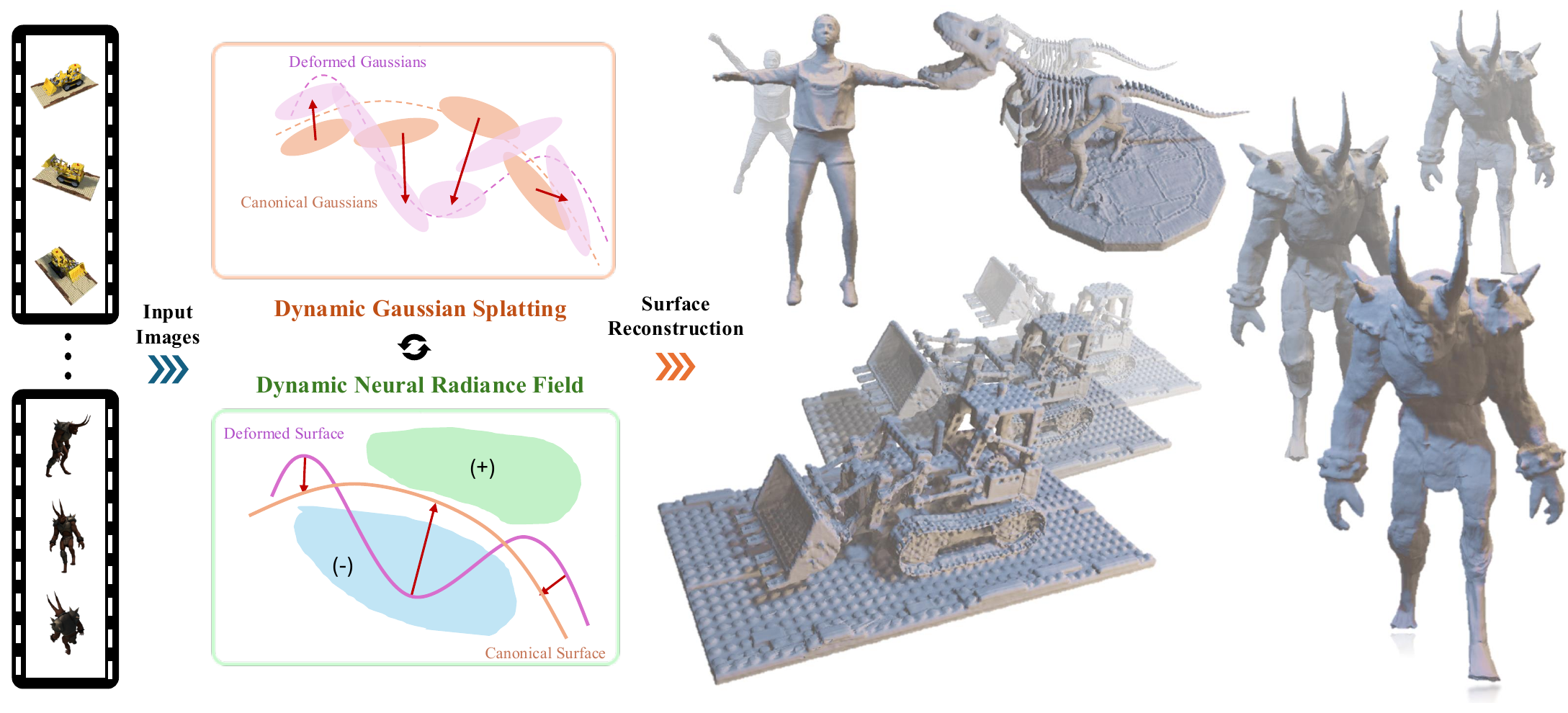}
\caption{Given a video sequence, DySurface obtains temporally consistent, high-quality meshes by bridging between explicit Gaussian primitives and implicit continuous surface functions.}
\label{teaser}
\end{figure}
\vspace{-0.1in}

\section{Introduction}
\label{sec:intro}
3D object reconstruction remains an important task in computer vision. Neural Radiance Fields (NeRF) \cite{nerf} and 3D Gaussian Splatting (3DGS) \cite{3dgs} have significantly advanced the field for static scenes, achieving high-fidelity rendering and accurate geometry. Recently, numerous works \cite{tineuvox, ndvg, rodynrf, nrnerf, suds} have extended these representations to dynamic scenes, learning object deformations through the rendering process.  

While prior methods effectively model temporal dynamics to achieve high photometric quality, they often fail to preserve fine-grained geometric details. NeRF-based approaches \cite{dnerf, tineuvox, hypernerf, nerfies, ffdnerf, v4d} exhibit relatively low rendering quality and limited capability in recovering high-frequency geometry. In contrast, Gaussian Splatting-based methods \cite{gags, 4dgs, dash, Grid4D, dynamic2dgs} achieve photorealistic rendering and maintain real-time synthesis speeds by explicitly modeling geometric deformations. However, direct visualization of optimized Gaussians reveals significant artifacts, such as floaters or primitives detached from accurate surface geometry. 

It is crucial to note that recovering explicit, deformable geometric surfaces is a fundamentally more demanding task than dynamic novel view synthesis (NVS). While dynamic NVS primarily requires photometric consistency across viewpoints, its surface reconstruction demands strict geometric and topological fidelity. Consequently, discrete representations optimized solely for view synthesis, such as dynamic 3DGS, fail to extract accurate meshes. These limitations necessitate a new paradigm that explicitly constructs discrete deformations with continuous surface regularizations.

Extracting meshes from pretrained deformable Gaussian Splatting models is non-trivial; the lack of explicit surface constraints often leads to geometric inconsistencies, resulting in noisy and blurred mesh reconstructions. DG-Mesh \cite{dgmesh} tackles the task of reconstructing surfaces from dynamic scenes. However, since it extracts meshes directly from discrete Gaussian Splatting via ad-hoc algorithms (e.g., Poisson reconstruction \cite{poisson}), it inherently lacks continuous surface constraints. This structural limitation results in blurry, fragmented, and inaccurate geometry under severe non-rigid deformations.

To address these challenges, we propose DySurface, a novel framework to extract highly accurate, temporally consistent deformable meshes from dynamic scenes. We tightly integrate explicit Gaussian primitives, implicit Signed Distance Functions (SDF), and explicit mesh representations into a unified pipeline. Specifically, our novel VoxGS-DSDF (Voxelized GS-driven Dynamic SDF) branch learns the continuous zero-level set of the object surface from the explicit geometric and appearance priors provided by the deformed Gaussians. Finally, we refine the extracted dynamic mesh by transferring the pretrained spatial transformation field from the GS branch, bridging the domain gap between the volumetric and explicit surface rendering. In summary, our main contributions are as follows:
\begin{itemize}
\item We propose DySurface, a tightly coupled framework that bridges the gap between discrete Gaussians, implicit continuous SDFs, and explicit meshes as mutually complementary representations.
\item We introduce a novel VoxGS-DSDF branch, which explicitly anchors the dynamic implicit surface using sparse voxel latents derived from deformed Gaussian splatting, effectively resolving geometric ambiguities.
\item Experiments demonstrate that DySurface achieves state-of-the-art geometric accuracies, reconstructing temporally consistent 4D meshes while maintaining rendering quality.
\end{itemize}

\section{Related Work}
\subsection{Dynamic View Synthesis}
Recent novel view synthesis (NVS) \cite{lumigraph, view, environment, lightfield, advances} has extended from static to dynamic scenes, where the primary challenge lies in accurately modeling object deformations. Among several works \cite{dnerf, tineuvox, hypernerf, kplanes, hexplane, neural-scene-flow, nerflow}, D-NeRF \cite{dnerf} introduced a canonical neural radiance field with a continuous backward deformation network. To overcome the slow convergence of original NeRFs and handle more extreme dynamic changes, subsequent works such as TiNeuVox \cite{tineuvox} and HyperNeRF \cite{hypernerf} proposed utilizing explicit spatiotemporal feature grids and mapping deformations into higher-dimensional spaces, respectively. Despite achieving high photometric quality, NeRF-based dynamic models struggle to capture fine geometric details due to their reliance on backward mapping \cite{ffdnerf}, learning non-smooth and discontinuous deformation fields. Furthermore, HexPlane \cite{hexplane} and K-Plane \cite{kplanes} project 4D spatial-temporal space to multiple 2D planes; thereby losing 3D features.

To overcome the limitations of implicit fields, recent literature \cite{4dgs, gags, voxelsplat, ex4dgs, gs4d, dynasurfgs} has extended the explicit 3DGS \cite{3dgs} framework to the deformable domain. 4D-GS \cite{4dgs} achieves real-time dynamic synthesis by seamlessly integrating temporal attributes into the 3DGS, modeling continuous motion via spatial-temporal HexPlanes. Similarly, GaGS \cite{gags} explicitly learns the non-rigid deformation of Gaussians, mapping with a sparse voxel grid, achieving high fidelity of rendering quality. While these GS-based methods establish the photometric rendering quality, they optimize strictly discrete, unconstrained primitives. Consequently, they are limited to view synthesis; without explicit surface regularizations, they fail to yield continuous, watertight geometric surfaces.

\subsection{Surface Reconstruction}
\noindent \textbf{Static Scenes.}
Traditional volume rendering lacks continuous geometric constraints, making accurate mesh extraction non-trivial. To address this, numerous works \cite{neus, neus2, volsdf, neuralangelo, regsdf, deepsdf, idr, mvsdf, probesdf} have incorporated SDF into the neural rendering pipeline. Specifically, NeuS \cite{neus} and VolSDF \cite{volsdf} successfully model solid surfaces by converting SDF values into density distributions for volume rendering. More recently, Neuralangelo \cite{neuralangelo} utilizes multi-resolution hash grids to recover high-frequency geometric details. However, these NeRF-based implicit models are constrained to static scenes; extending them to dynamic environments while preserving highly accurate geometric details remains a formidable challenge due to the complex, non-rigid topological variations over time.
While 3DGS \cite{3dgs} offers a highly efficient representation, 3DGS optimizes discrete primitives driven purely by a photometric loss, thus lacking continuous surface constraints. Recent literature \cite{sugar, gs2mesh, 3dgs-to-pc, 2dgs, pgsr, neusg, gsdf, 3dgsr} has attempted to extract surfaces from this discrete representation. SuGaR \cite{sugar} introduces regularization terms to align Gaussians with the underlying geometry. Alternatively, GS2Mesh \cite{gs2mesh} avoids direct extraction by leveraging a stereo-matching network to generate depth maps from novel views, while 3DGS-to-PC \cite{3dgs-to-pc} proposes probabilistically sampling dense point clouds from 3DGS. Despite these efforts, these methods rely on ad-hoc post-processing (e.g., Poisson reconstruction) or external 2D priors, thus struggle to maintain a globally consistent zero-level set, often yielding fragmented reconstructions that lack topological coherence, particularly in regions where the discrete Gaussian representation deviates from the actual surface geometry.

\setlength{\columnsep}{11pt}
\begin{wrapfigure}{r}{0.46\textwidth}
  \begin{center}
  \advance\leftskip+0mm
  \renewcommand{\captionlabelfont}{\footnotesize}
    \vspace{-0.12in}  
  \includegraphics[width=0.45\textwidth]{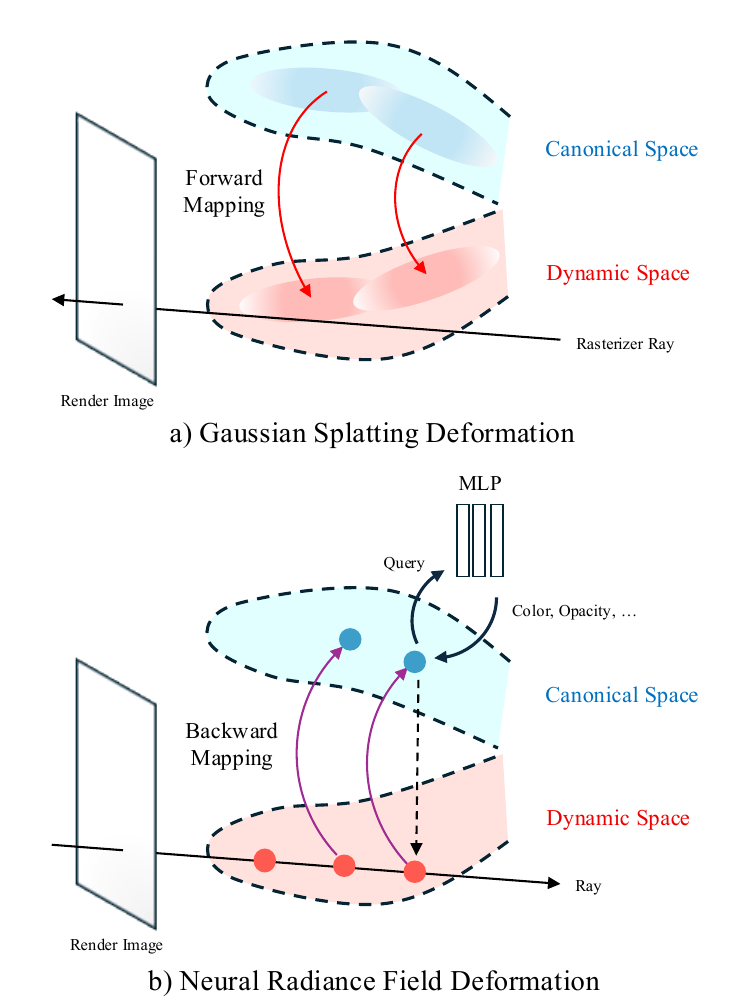}
    \vspace{-0.05in} 
     \caption{Illustration of forward mapping from Gaussian Splatting and backward mapping from neural radiance field.}
    \label{fig:pre}
    \vspace{-0.2in} 
  \end{center}
\end{wrapfigure}
To harness both the explicit 3DGS and the implicit geometric fidelity of SDF, recent work \cite{gsdf, discretizedsdf, 3dgsr} has explored hybrid architectures. Notably, GSDF \cite{gsdf} introduces a dual-branch framework that jointly optimizes an explicit 3DGS branch and an implicit SDF branch, employing mutual spatial guidance to enhance both view synthesis and surface reconstruction. While achieving state-of-the-art 3D geometry results, a major limitation of GSDF is its confinement to static scenes. The mutual guidance mechanism assumes a static coordinate space and lacks temporal deformation modeling. Therefore, it fails to reconstruct non-rigid, deformable changes of the object over time.

\noindent \textbf{Dynamic Scenes.}
While surface reconstruction for static scenes has been extensively studied, it remains under-explored in dynamic environments. For instance, DG-Mesh \cite{dgmesh} attempts to recover the surfaces of deformable objects by extracting a preliminary mesh from 3D Gaussian Splatting via Poisson surface reconstruction, followed by a refinement stage. However, applying Poisson reconstruction to unconstrained, discrete Gaussians intrinsically fails to preserve geometric accuracy. D-2DGS \cite{dynamic2dgs} also extracts a deformable mesh from 2DGS \cite{2dgs}; frame-by-frame extraction does not guarantee geometric consistency. NeRF-based methods \cite{dnerf, tineuvox, hypernerf} extract surfaces by thresholding volume density; they still suffer from severe geometric ambiguity and lack of fidelity due to the absence of explicit surface regularization. In contrast, our approach achieves state-of-the-art mesh extraction accuracy by synergizing the explicit and efficient geometric deformation learned by 3D Gaussians with precise SDF fields optimized via volume rendering.

\vspace{-0.2cm}
\section{Preliminary}
\label{sec:preliminary}
\noindent \textbf{Deformation Formulations: Forward vs. Backward Mapping.} 
In modeling dynamic scenes, 3D Gaussian Splatting (3DGS) \cite{3dgs} and Neural Radiance Fields (NeRF) \cite{nerf} adopt opposite deformation directions. 
As illustrated in Fig. \ref{fig:pre}, dynamic 3DGS employs a forward mapping framework.
It explicitly transforms the geometric attributes (e.g., positions, rotations) of discrete Gaussians from a canonical space to the dynamic observation space ($T_{canonical \rightarrow dynamic}$). 
Conversely, NeRF relies on a backward mapping approach, since it synthesizes images by casting rays into the dynamic observation space. It warps the sampled 3D points along the rays back to a canonical space ($T_{dynamic \rightarrow canonical}$) to query continuous volume density and radiance.

\begin{figure*}[t]
  \centering
  \includegraphics[width=13.3cm]{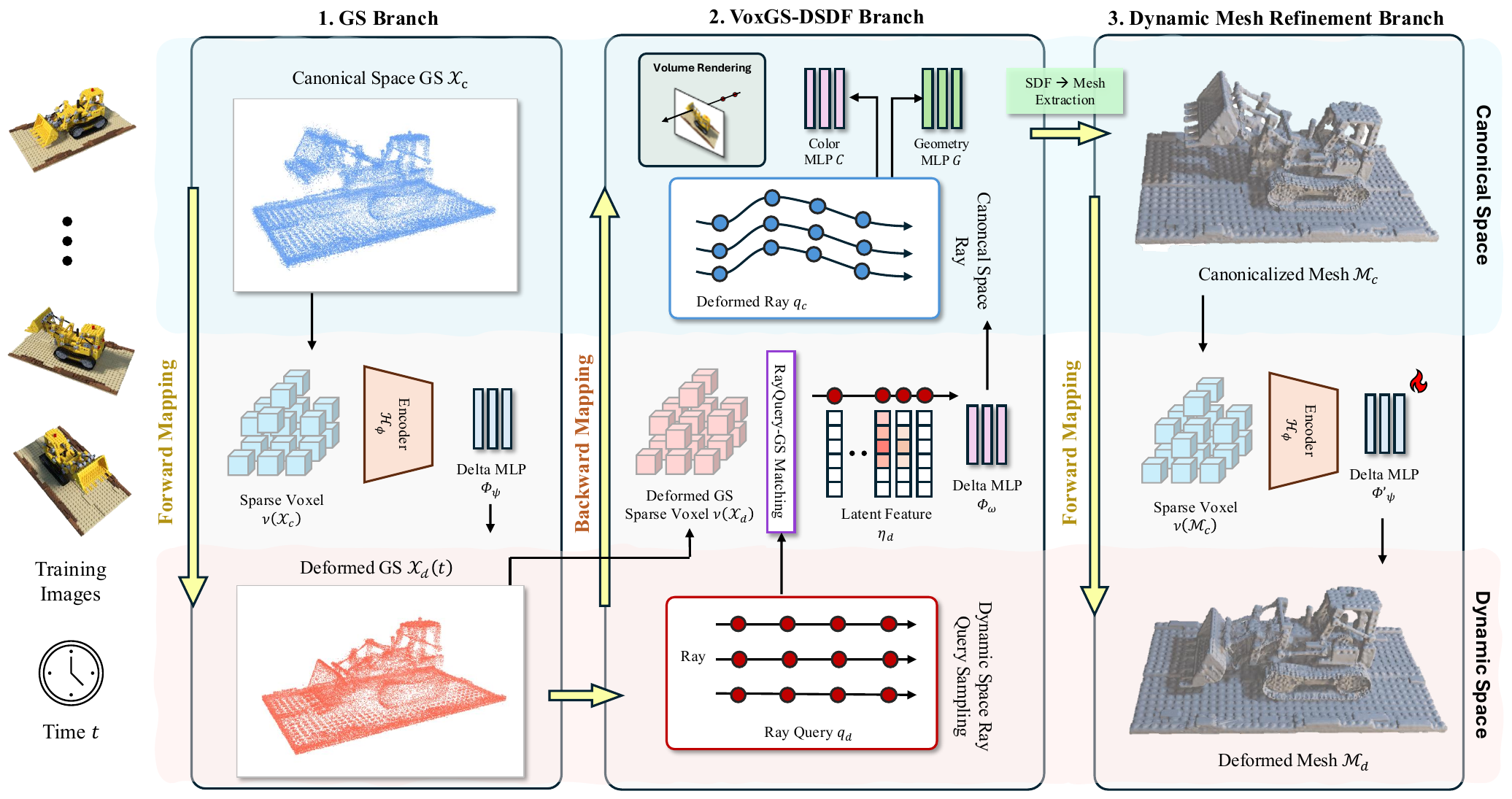}
  \caption{\textbf{Architecture of DySurface.} Our framework consists of three branches, learned in a stage-wise manner. (1) GS branch deforms canonical 3DGS into dynamic space using a transform field. (2) VoxGS-DSDF branch anchors the implicit SDF field to the deformed GS voxels through RayQuery-GS matching, resolving the forward-backward mapping gap while volume rendering. (3) The final branch extracts the canonical surface from the learned SDF and refines the transform field.}
  \label{fig:architecture}
\end{figure*}

This inherent distinction arises from their respective rendering processes: NeRF queries a continuous static field via backward mapping, whereas 3DGS relies on the explicit forward rasterization of moving discrete primitives. Consequently, directly applying SDF-based surface regularization to dynamic Gaussian Splatting is a non-trivial challenge, as it requires resolving the structural conflict between forward-mapped dynamics and backward-mapped implicit surfaces.

\section{Method}

\label{SVDNBranch}
In this section, we introduce DySurface, a method for reconstructing a detailed deformable mesh from a dynamic scene. Our method is constructed with three stages: i) the Gaussian Splatting (GS) branch (Sec. \ref{sec:GSBranch}), ii) the VoxGS-DSDF branch (Sec. \ref{sec:VoxGSDF}), and iii) the Dynamic-Mesh Refinement branch (Sec. \ref{sec:DMRBranch}). The overall framework is illustrated in Fig. \ref{fig:architecture}

\subsection{Gaussian Splatting Branch}
\label{sec:GSBranch}
Following GaGS \cite{gags}, our Gaussian Splatting (GS) branch explicitly models the temporal deformation of canonical 3D Gaussian primitives $\mathcal{X}_c$, parameterized by mean position $\mu_c \in \mathbb{R}^3$, scale $s_c \in \mathbb{R}^3$, and rotation $r_c \in \mathbb{R}^4$, along with time-invariant opacity $\alpha$ and SH coefficients.

To model non-rigid dynamics, we define a forward transformation field $\Delta \overrightarrow{q}_{fwd}$. We first discretize the canonical gaussians $\mathcal{X}_c$ into a sparse voxel grid $\nu(\mathcal{X}_c)$ to extract a global feature volume via a 3D sparse convolutional encoder $\mathcal{H}_\phi$. For any coordinate $x$, its queried global latent $\mathbf{f}_{global}(x)$ is concatenated with a point-wise local feature $\mathbf{f}_{local}(x)$. An MLP decoder $\Phi_\psi$ then predicts the time-dependent residuals, formulating the transformation field $\mathcal{T}_\theta$ (parameterized by $\theta = \{\phi, \psi\}$) as:
\begin{equation}
\mathcal{T}_\theta(x, t) = \Phi_\psi(\text{Concat}(\mathbf{f}_{global}(x), \mathbf{f}_{local}(x)), t)
\end{equation}

Applying these residuals yields the explicitly deformed attributes for the $k$-th Gaussian at time $t$. We denote the per-component output of the transformation field as $\mathcal{T}_\theta(\mu_{c,k}, t) = (\Delta\mu_k, \Delta s_k, \Delta r_k)$:
\begin{equation}
\mu_{d,k}(t) = \mu_{c,k} + \Delta\mu_k, \quad
s_{d,k}(t)  = s_{c,k}  + \Delta s_k,  \quad
r_{d,k}(t)  = \frac{r_{c,k} + \Delta r_k}{\|r_{c,k} + \Delta r_k\|}
\end{equation}

The resulting deformed primitives, denoted as $\mathcal{X}_d$, synthesize the novel view at time $t$ and serve as the geometric prior for the VoxGS-DSDF branch.

\subsection{Voxelized GS-driven Dynamic SDF Branch}
\label{sec:VoxGSDF}

The Voxelized GS-driven Dynamic SDF (VoxGS-DSDF) branch learns a high-fidelity continuous SDF in the canonical space. To overcome the lack of geometric guidance in implicit methods, we leverage the deformed Gaussians $\mathcal{X}_d(t) = \{x_d^i(t)\}_{i=1}^{N_G}$ from the GS branch to construct a dynamic sparse voxel grid, anchoring the continuous volumetric field to explicit geometric priors, where $x_d^i(t)$ denotes the $i$-th deformed Gaussian primitive.

As detailed in Sec. \ref{sec:preliminary}, the rendering begins by sampling points $q_d$ along rays in the dynamic space. To extract spatial context for each query point $q_d$, we perform a K-Nearest Neighbors (KNN) search to identify the explicitly deformed Gaussian sparse voxels $\nu(\mathcal{X}_d(t))$. Let $\mathcal{N}_{K}(q_d; \nu(\mathcal{X}_d(t)))$ denote the initial set of the $K$-nearest voxel neighbors of $q_d$. To filter out irrelevant features, we apply a spatial distance threshold $\tau$, retaining only the valid subset of neighbors
$\mathcal{N}'(q_d; \nu(\mathcal{X}_d(t))) =
\{ k \in \mathcal{N}_K(q_d; \nu(\mathcal{X}_d(t))) \mid \|v_k - q_d\|_2^2 < \tau \}$,
where $v_k$ is the center position of the $k$-th neighbor in $\nu(\mathcal{X}_d(t))$. From these $K' = |\mathcal{N}'(q_d; \nu(\mathcal{X}_d(t)))|$ valid neighbors, we extract geometric features $\mathbf{f}_{geo,k}$ from the deformed mean positions and appearance features $\mathbf{f}_{app,k}$ from the embedded color attributes. We then concatenate these features with the positional encoding $\gamma(q_d)$ to construct the spatial conditioning vector $\eta_d$.
\begin{equation}
\eta_d = \text{Concat}\left(
\gamma(q_d),
\frac{1}{K'}\sum_{k \in \mathcal{N}'(q_d; \nu(\mathcal{X}_d(t)))} \mathbf{f}_{geo, k},
\frac{1}{K'}\sum_{k \in \mathcal{N}'(q_d; \nu(\mathcal{X}_d(t)))} \mathbf{f}_{app, k}
\right)
\end{equation}

% To extract spatial context for each query point $q_d$, we perform K-Nearest Neighbors (KNN) search $\mathbb{i}$ to query the attributes of explicitly deformed Gaussians. To filter irrelevant features, we apply a distance threshold $\tau$, retaining only neighbors satisfying $\|\mu_k - q_d\|^2 < \tau$. From these $K'$ valid neighbors, we extract and average their geometric features $\mathbf{f}_{geo}$ and appearance features $\mathbf{f}_{app}$ (derived from SH coefficients). The complete latent representation $\eta_d$ for $q_d$ is constructed by concatenating its harmonic positional encoding $\gamma(x_d)$ with these aggregated features:

Conditioned on a temporal embedding $t$, an MLP $\Phi_{\omega}$ takes latent $\eta_d$ to predict the backward deformation $\Delta \overleftarrow{q}_{bwd}$, mapping the dynamic point back to the canonical space: $q_c = q_d + \Phi_{\omega}(\eta_d, t)$.

The mapped canonical coordinate $q_c$ is fed into a geometry network $G$ to predict the SDF value $s_i = G(q_{c, i})$ and geometric features, which are then passed to a color network $C$ for view-dependent radiance $c_i$. The SDF values are converted into discrete alpha values $\alpha_i$ using a learnable logistic distribution $\mathcal{G}(x) = (1 + e^{\beta x})^{-1}$ with parameter $\beta$ controlling surface sharpness:
\begin{equation}
\alpha_i = \max\left(\frac{\mathcal{G}(s_i) - \mathcal{G}(s_{i+1})}{\mathcal{G}(s_i)}, 0\right)
\end{equation}
Finally, the predicted pixel color $\hat{C}(\mathbf{r})$ is synthesized via standard volumetric integration, accumulating radiance and opacity of the number of $N$ sampled points along the ray:
\begin{equation}
\hat{C}(\mathbf{r}) = \sum_{i=1}^{N} T_i \alpha_i c_i, \quad \text{where} \quad T_i = \prod_{j=1}^{i-1} (1 - \alpha_j)
\end{equation}
where $T_i$ is the accumulated transmittance, ensuring the rendering respects the implicit zero-level set.

\subsection{Dynamic Mesh Refinement Branch}
\label{sec:DMRBranch}
Upon convergence of the VoxGS-DSDF branch, we extract a high-fidelity canonical mesh $\mathcal{M}_c$ through the mesh extraction algorithm \cite{marchingcube} to the zero-level set of the SDF. As implicit SDF field and explicit GS primitives are anchored and optimized within the same canonical space, the extracted mesh vertices $v_c \in \mathcal{M}_c$ are inherently aligned with the canonical Gaussians.

To articulate a static canonical mesh $v_c$ across the temporal domain, we can seamlessly reuse the forward transformation field already learned in the GS branch. As explicit discrete primitives, 3D Gaussians excel at tracking fast and complex non-rigid dynamics during the initial optimization. We transfer this pre-trained explicit motion prior to the mesh surface. The dynamic vertices $v_d(t)$ at any given time step $t$ are obtained by querying the spatial deformation component of the transformation field $\mathcal{T}_{\theta'}$ initialized from $\mathcal{T}_\theta$, thus is defined as $v_d(t) = v_c + [\mathcal{T}_{\theta'}(v_c, t)]_\mu$.

% Using a differentiable mesh rasterizer, we optimize the shared transformation network directly on the extracted dynamic mesh. This stage yields a final dynamic mesh that retains the underlying topology of the implicit field while inheriting the motion dynamics of the explicit Gaussians.

\subsection{Learning Objective}

\noindent \textbf{Gaussian Splatting Branch.}
In the first stage, we optimize the canonical 3DGS and the forward transformation field $\mathcal{T}_\theta$. The overall loss is:
\begin{equation}
\mathcal{L}_{GS} = \mathcal{L}_{rgb} + \lambda_{m} \mathcal{L}_{mask} + \lambda^{gs}_{reg} \mathcal{L}_{reg}^{gs}
\end{equation}

where $\mathcal{L}_{rgb}$ combines $\mathcal{L}_1$ and SSIM, $\mathcal{L}_{mask}$ is the opacity $\mathcal{L}_1$ error, and $\mathcal{L}_{reg}^{gs} = \|\Delta \mu\|_1$ encourages minimal spatial deformations.

\noindent \textbf{VoxGS-DSDF Branch.}
We train the SDF network ($C, G$) and the backward mapping field $\Phi_{\omega}$. Alongside the rendering losses ($\mathcal{L}_{rgb}^{sdf}$ and $\mathcal{L}_{mask}^{sdf}$), we propose constraints to bridge explicit and implicit representations.

\textit{Cycle Consistency Loss.} To ensure bijection between the forward ($\Delta \overrightarrow{q}_{fwd}$ and backward ($\Delta \overleftarrow{q}_{bwd}$) deformations, we minimize their sum:
\begin{equation} 
\mathcal{L}_{cycle} = \|\Delta \overleftarrow{q}_{fwd} + \Delta \overrightarrow{q}_{bwd}\|_1
\end{equation}

\textit{SDF-GS Anchoring Loss.} To guide the continuous zero-level set with discrete primitives, we penalize their divergence using a BCE loss between absolute SDF values and the K-nearest canonical Gaussian centers. Let ${d_{NN}}$ denote the mean distance from $q_c$ to centers in $\mathcal{N}_K$.
\begin{equation} 
\mathcal{L}_{SDF-GS} = \text{BCE}(|s(q_c)| > \epsilon, d_{NN}(q_c; \mathcal{X}_c) > \epsilon)
\end{equation}

We distill geometric details using depth and normal maps from the GS branch ($\mathcal{L}_{distill}$), and apply Eikonal ($\mathcal{L}_{eik} = (\|\nabla s(q_c)\|_2 - 1)^2$) and temporal smoothing ($\mathcal{L}_{smooth}$) regularizations to the total SDF loss $\mathcal{L}_{SDF}$. The temporal smoothing term $\mathcal{L}_{smooth}$ is described in the supplementary.
\begin{equation} 
\mathcal{L}_{SDF} = \mathcal{L}_{rgb}^{sdf} + \lambda_{c} \mathcal{L}_{cycle} + \lambda_{SG} \mathcal{L}_{SDF-GS} + \lambda_{dist} \mathcal{L}_{distill} + \lambda_{e} \mathcal{L}_{eik} + \lambda_{s} \mathcal{L}_{smooth}
\end{equation}

\noindent \textbf{Dynamic Mesh Refinement.} 
In the final fine-tuning stage, we optimize the refined forward mapping field network $\mathcal{T'}_\theta$. The refinement loss is defined as:
\begin{equation} 
\mathcal{L}_{Mesh} = \mathcal{L}_{rgb}^{mesh} + \lambda^{mesh}_{m} \mathcal{L}_{mask}^{mesh} + \lambda_{lap} \mathcal{L}_{lap} + \lambda^{mesh}_{reg} \mathcal{L}_{reg}^{mesh},
\end{equation}
where $ \mathcal{L}_{reg}^{mesh} = \|v_d - v_c\|_1$. For stable geometry, Laplacian smoothing ($\mathcal{L}_{lap}$) is applied to vertices with the lowest 25\% curvature errors. The $\mathcal{L}_{reg}^{mesh}$ regularizes the minimal deformation of the object.

\section{Experiments}
\subsection{Datasets}
The D-NeRF \cite{dnerf} dataset provides synthetic multi-view renderings of dynamic objects; however, it only supplies photometric images as ground truth and lacks evaluation of 3D geometry metrics. Since the original dataset does not officially release the 3D sequence meshes, we curated a novel 3D benchmark by sourcing the publicly accessible original assets. Specifically, we perform experiments on three scenes: \textit{Lego}, \textit{Hell Warrior}, and \textit{T-Rex}. For each scene, we extracted the ground-truth 3D meshes at three distinct temporal snapshots: $t \in \{0.0, 0.5, 1.0\}$. This allows us to quantitatively measure the geometric fidelity and accuracy of the reconstructed surfaces over time using pure 3D metrics. Qualitative results on the remaining scenes can be found in the supplementary material. Tab. \ref{tab:quantitative} uses three snapshots, while we conduct a dense 60-frame evaluation in the supplementary.

To further validate the robustness and generalization of our proposed framework, we conduct experiments on the DG-Mesh \cite{dgmesh} and Nerfies \cite{nerfies} datasets. The DG-Mesh dataset features diverse deforming objects rendered from 360-degree viewpoints and provides ground-truth meshes. This allows us to evaluate the 3D geometric metrics using the same criteria applied to our D-NeRF benchmark. Finally, the Nerfies dataset consists of in-the-wild video captures of complex, non-rigidly deforming subjects recorded by a moving camera. Since high-fidelity 3D geometric ground truth is not available for such real-world captures, our evaluation on this dataset focuses primarily on qualitative mesh and quantitative NVS metrics, following the prior work \cite{dgmesh}. There are additional real-world datasets; we experiment with DyCheck \cite{dycheck} in the supplementary material.

% The Nerfies dataset consists of stereo-view video captures of complex, causal animals recorded by a moving camera. Since obtaining high-fidelity geometric ground truths is intractable for in-the-wild video captures, our evaluation on the Nerfies dataset focuses on qualitative mesh and quantitative novel view synthesis metrics, following prior work \cite{dgmesh}.
% To further validate the robustness and generalizability of our proposed model, we conduct experiments on the widely used DG-Mesh \cite{dgmesh} and Nerfies \cite{nerfies} dataset for real-world scenarios. The DGMesh datasets consists diverse deforming objects rendered from 360 views with given ground-truth mesh. We measure 3D geometric metrics with the same criteria to the D-NeRF
% three scenes exhibiting complex deformations were available:

\begin{table*}[t]
\centering
\caption{\textbf{Quantitative comparisons on the D-NeRF \cite{dnerf} dataset.} We evaluate both 3D geometric accuracy and photometric rendering quality. The best, second-best, and third-best results are highlighted in \colorbox{rred}{red}, \colorbox{oorange}{orange}, and \colorbox{yyellow}{yellow}, respectively. }
\label{tab:quantitative}
\begin{adjustbox}{scale=0.82}
\begin{tabular}{@{}lccc|ccc@{}}
\toprule
Scene          & \:\: Methods  \:\:  & \: vIoU (↑) \: & \: CD (↓) \: & \: PSNR (↑)  & SSIM (↑)  &  \: LPIPS (↓) \: \\ \midrule
\multirow{6}{*}{\textit{``Lego''}}    
& \: TiNeuVox \cite{tineuvox} \: & 0.1080 & 0.0174 & 25.17 & 0.9217 & 0.0689 \\                                        
& NeuS2 \cite{neus2}  & 0.2578 & 0.0552 & 22.88 & 0.8945 & 0.0716 \\ 
& 4DGS \cite{4dgs}  & 0.2154 & 0.0180 & \cellcolor{oorange}{25.40} & \cellcolor{yyellow}{0.9434} & \cellcolor{yyellow}{0.0377} \\
& GaGS  \cite{gags}   & \cellcolor{yyellow}{0.2309} & \cellcolor{yyellow}{0.0155} & \cellcolor{rred}{25.44} & \cellcolor{rred}{0.9474} & \cellcolor{oorange}{0.0329} \\ \cmidrule{2-7}
& DG-Mesh \cite{dgmesh}  & \cellcolor{oorange}{0.2655} & 0.0236 & 21.29 & 0.8382 & 0.1590 \\
& D-2DGS \cite{dynamic2dgs}  & 0.2138 &  \cellcolor{oorange}{0.0135}  & 23.29 & 0.8872 & 0.1121 \\
& Ours     & \cellcolor{rred}{0.2988} & \cellcolor{rred}{0.0133} & \cellcolor{oorange}{25.40} & \cellcolor{oorange}{0.9439} & \cellcolor{rred}{0.0245} \\ \midrule
\multirow{6}{*}{\textit{``T-Rex''}} 
& TiNeuVox & 0.4232 & 0.0071 & 32.77 & 0.9783 & 0.0307 \\
& NeuS2  &  -   & -   & -  &  -  & -  \\ 
& 4DGS     & 0.2702 & 0.0710 & \cellcolor{yyellow}{33.39} & \cellcolor{yyellow}{0.9869} & \cellcolor{yyellow}{0.0130}  \\
& GaGS     & 0.2530 & 0.0069 & \cellcolor{rred}{39.02} & \cellcolor{rred}{0.9952} & \cellcolor{rred}{0.0052} \\ \cmidrule{2-7}
& DG-Mesh  & \cellcolor{oorange}{0.5736} & \cellcolor{rred}{0.0044} & 28.95 & 0.9590 & 0.0651 \\
& D-2DGS   & \cellcolor{yyellow}{0.4997} & \cellcolor{yyellow}{0.0054} & 28.68 & 0.9668 & 0.0434 \\
& Ours     & \cellcolor{rred}{0.5783} & \cellcolor{oorange}{0.0050} & \cellcolor{oorange}{36.75} & \cellcolor{oorange}{0.9932} & \cellcolor{oorange}{0.0069} \\ \midrule
\multirow{6}{*}{\textit{``HellWarrior''}}
& TiNeuVox & 0.2468 & 0.0252 & \cellcolor{yyellow}{28.20} & 0.9661 & 0.0631 \\
& NeuS2  & 0.2548 & 0.0326 & 18.02 & 0.8977 & 0.1064 \\ 
& 4DGS     & 0.2296 & 0.0221 & 28.12 & \cellcolor{yyellow}{0.9730} & \cellcolor{yyellow}{0.0276} \\
& GaGS     & 0.2427 & 0.0215 & \cellcolor{rred}{32.27} & \cellcolor{rred}{0.9835} & \cellcolor{rred}{0.0164} \\ \cmidrule{2-7}
& DG-Mesh  & \cellcolor{oorange}{0.3068} & \cellcolor{yyellow}{0.0198} & 25.46 & 0.9590 & 0.0841 \\
& D-2DGS & \cellcolor{rred}{0.3079} & \cellcolor{oorange}{0.0132} & 25.50 & 0.9583 & 0.0488 \\
& Ours     & \cellcolor{yyellow}{0.3024} & \cellcolor{rred}{0.0124} & \cellcolor{oorange}{31.14} & \cellcolor{rred}{0.9835} & \cellcolor{oorange}{0.0169} \\ \midrule
\multirow{6}{*}{Average}
& TiNeuVox & 0.2593 & 0.0165 & 28.71 & 0.9382 & 0.0673 \\
& NeuS2    &  -  &   -     &   -    &   -   &   -   \\  
& 4DGS     & 0.2384 & 0.0370 & \cellcolor{yyellow}{28.97} & \cellcolor{yyellow}{0.9677} & \cellcolor{yyellow}{0.0261} \\
& GaGS     & 0.2422 & 0.0146 & \cellcolor{rred}{32.24} & \cellcolor{rred}{0.9753} & \cellcolor{oorange}{0.0181} \\ \cmidrule{2-7}
& DG-Mesh  & \cellcolor{oorange}{0.3814} & \cellcolor{yyellow}{0.0137} & 25.23 & 0.9186 & 0.1027 \\
& D-2DGS  & \cellcolor{yyellow}{0.3426} & \cellcolor{oorange}{0.0111} & 25.82 & 0.9374 & 0.0681 \\
& Ours     & \cellcolor{rred}{0.3932} & \cellcolor{rred}{0.0102} & \cellcolor{oorange}{31.09} & \cellcolor{oorange}{0.9735} & \cellcolor{rred}{0.0161} \\ \bottomrule
\end{tabular}
\end{adjustbox}
\vspace{-0.3cm}
\end{table*}

\begin{figure*}[t]
  \centering
  \includegraphics[width=\textwidth]{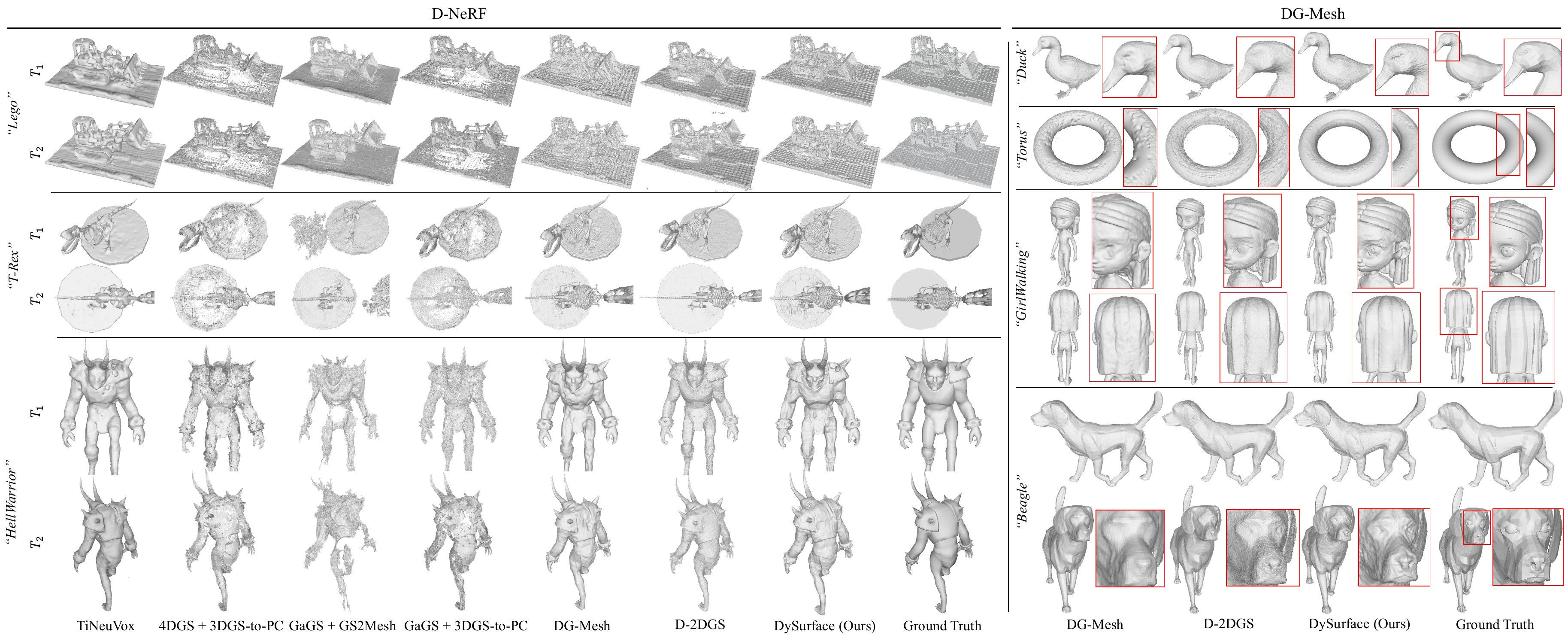}
  \caption{\textbf{Qualitative results of D-NeRF \cite{dnerf} and DG-Mesh \cite{dgmesh} dataset.} We visualize the results at two different time steps from distinct viewpoints. Please zoom to check details.}
  \label{fig:qualitative}
  \vspace{-0.4cm}
\end{figure*}

\begin{figure}[h]
    \begin{minipage}[t]{0.45\linewidth}
        \vspace{0pt}
        \centering
        \includegraphics[width=\linewidth]{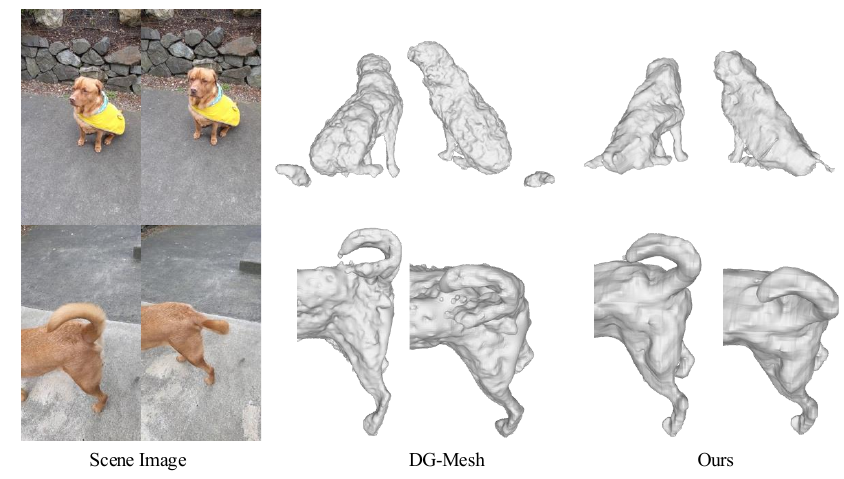} 
        \caption{\textbf{Qualitative results on Nerfies \cite{nerfies}.}}
        \label{fig:quantitative2}
    \end{minipage}
    \hfill
    \begin{minipage}[t]{0.5\linewidth}
        \vspace{0pt} 
        \centering
        \captionof{table}{\textbf{Quantitative results on Nerfies \cite{nerfies} dataset.} Our method maintains better rendering quality than DG-Mesh.}
        \label{tab:quantitative2}
        
            \begin{adjustbox}{scale=0.94}
            \begin{tabular}{@{}l|cc@{}}
            \toprule
            Methods          & \: DG-Mesh \cite{dgmesh} \: & \: Ours \: \\ \midrule
            
             PSNR($\uparrow$)     & 27.23   &  \textbf{27.91}     \\
             SSIM ($\uparrow$)    & 0.9126  &  \textbf{0.9201}    \\ 
             LPIPS ($\downarrow$) & 0.0927  &  \textbf{0.0815}    \\ \bottomrule
            \end{tabular}
            \end{adjustbox}
        
    \end{minipage}
\vspace{-0.5cm}
\end{figure}

\subsection{Experimental Setup}
\noindent\textbf{Baselines.} We evaluate our method against state-of-the-art methods \cite{tineuvox, gags, 4dgs, dgmesh} across dynamic scene representations and surface extraction. Since 3DGS-based methods \cite{gags, 4dgs} rely on discrete primitives without continuous surface modeling, we establish explicit baselines by applying recent mesh extraction techniques, specifically GS2Mesh \cite{gs2mesh} and 3DGS-to-PC \cite{3dgs-to-pc} to their outputs. As 3DGS-to-PC, built upon SuGaR \cite{sugar}, consistently yields better accuracy than GS2Mesh, we report its results in the main paper and defer GS2Mesh comparisons to the supplementary. Furthermore, we directly compare our approach against DG-Mesh \cite{dgmesh}, D-2DGS \cite{dynamic2dgs}, a dynamic mesh reconstruction framework, and TiNeuVox \cite{tineuvox}, a NeRF-based dynamic volume rendering method.
NeuS2 \cite{neus2} is evaluated under an adapted monocular setting using consecutive temporal frames as pseudo-multiview observations.

\noindent\textbf{Evaluation Metrics.}
To measure 3D geometric accuracy, we report Volumetric Intersection over Union (vIoU) and Chamfer Distance (CD) to assess the mesh topology. All 3D reconstruction metrics are reported with the average score across representative temporal states ($t \in \{0.0, 0.5, 1.0\}$). For the photometric quality of NVS, we employ PSNR, Structural Similarity Index (SSIM) \cite{ssim}, and Learned Perceptual Image Patch Similarity (LPIPS) \cite{lpips}.

\begin{figure*}[t]
\vspace{-0.2cm}
  \centering
  \includegraphics[width=\textwidth]{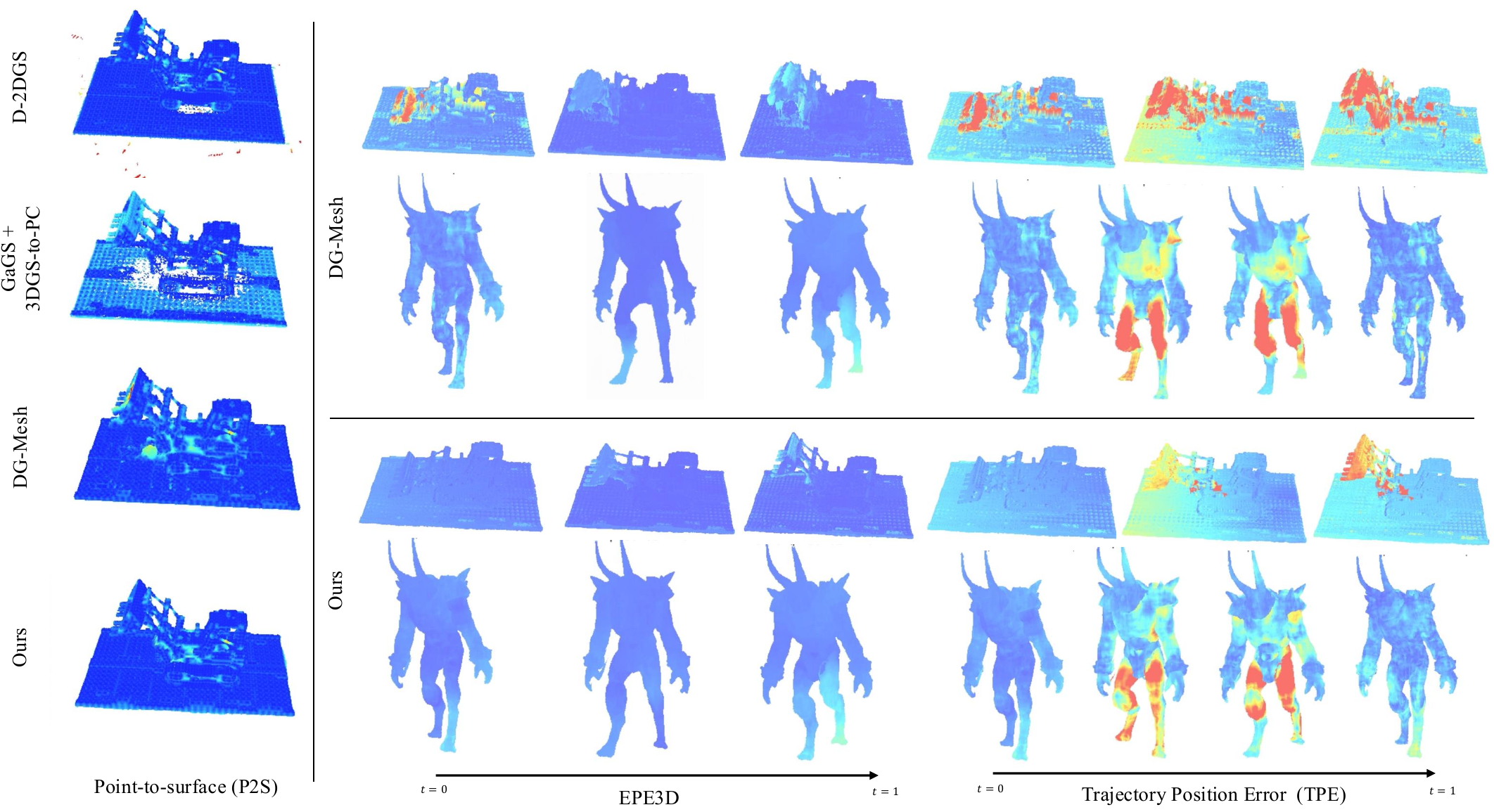}
  \caption{\textbf{Qualitative visualization of temporal consistency with a heatmap.} We show the heatmap rendered image of P2S, EPE3D, and TPE metrics.}
  \label{fig:tce}
  \vspace{-0.2cm}
\end{figure*}

We report temporal consistency performance with 3D End-point error (EPE-3D), Flow Angular Error, Trajectory Normal Angular Error (TNAE), and Trajectory Position Error (TPE) metrics. We sample surface points on the canonical mesh and track their predicted and ground-truth trajectories. Note that the per-frame reconstruction baselines \cite{dynamic2dgs, gags, 4dgs} reconstruct their mesh independently per time step. As they do not preserve consistent vertex correspondences across frames, vertex-level trajectory metrics are not directly defined. We therefore compare with DG-Mesh, which maintains a fixed-topology deformable mesh in Tab. \ref{tab:tce} and Fig. \ref{fig:tce}.

\noindent\textbf{Implementation Details.} We implement our framework using PyTorch \cite{pytorch} and optimize it on a single NVIDIA RTX 4090 GPU (24GB VRAM). Each stage optimization takes approximately 2 hours for Stage 1, 2 hours for Stage 2, and 1 hour for Stage 3 mesh refinement.
%The total optimization time is approximately 5 hours per dynamic scene. Specific details of the runtime can be found in the supplementary.

\subsection{Results}
\noindent\textbf{Quantitative Results.}
Tab. \ref{tab:quantitative} presents the quantitative comparisons on the D-NeRF \cite{dnerf}, Tab. \ref{tab:dgmesh_table} on the DG-Mesh \cite{dgmesh} dataset, and Tab. \ref{tab:quantitative2} on the Nerfies \cite{nerfies} dataset. Our method achieves state-of-the-art performance on most geometry metrics, maintaining high rendering quality. Notably, our method outperforms the mesh reconstruction baseline, DG-Mesh \cite{dgmesh} and D-2DGS \cite{dynamic2dgs}, not only in geometric fidelity but also by a large margin in photometric quality. Note that NeuS2 \cite{neus2} fails to converge in the \textit{T-Rex} scene.

\begin{wraptable}{r}{0.58\textwidth} % 'l'은 표를 왼쪽에 배치, '0.5\textwidth'는 표가 차지할 가로 너비 비율
  \centering
  \advance\leftskip-2mm
  \caption{\textbf{Quantitative results on DG-Mesh \cite{dgmesh} dataset.}}
  \label{tab:dgmesh_table}
  \begin{adjustbox}{scale=0.92}
  \centering
  \begin{tabular}{lcc|cc}
    \toprule
                              & vIoU($\uparrow$) & CD(↓) & PSNR($\uparrow$)  & LPIPS(↓)   \\
    \midrule
    D-2DGS \cite{dynamic2dgs} & 0.6147 & \underline{0.0084} & 24.84 & 0.0989  \\
    DG-Mesh \cite{dgmesh}     & \textbf{0.7602} & 0.0090 & \underline{30.45} & \underline{0.0554}   \\
    Ours                      & \underline{0.7571} & \textbf{0.0078} & \textbf{30.92} & \textbf{0.0401}  \\
    \bottomrule
  \end{tabular}
  \end{adjustbox}
\end{wraptable}

Our method achieves competitive photometric rendering performance against purely NVS-optimized methods such as GaGS \cite{gags} and 4DGS \cite{4dgs}. Crucially, while these NVS-centric approaches sacrifice the 3D geometry to minimize 2D projection errors, our framework simultaneously maintains high geometric accuracy. This proves that our architecture successfully enforces strict, continuous surface constraints without degrading rendering details.
% closely matching the visual fidelity 
% , reconstructing an accurate surface across time.

Tab. \ref{tab:tce} shows the evaluation results of temporal consistency. As shown in Tab. \ref{tab:tce}, our proposed method improves all four metrics, with particularly clear gains in TNAE and trajectory position error, supporting improved normal and trajectory stability. Our transform field learned in both explicit representations (GS and SDF Mesh) enables effective capture of surface flow with plausible geometries. These quantitative results demonstrate that our framework achieves accurate tracking stability with higher normal consistency.

\begin{table*}[h]
\centering
\caption{\textbf{Quantitative Comparison with Temporal and Trajectory Flow Metrics.}}
\label{tab:tce}
\begin{adjustbox}{scale=0.86}
\begin{tabular}{@{}c|cccc@{}}
\toprule
Method & EPE3D $\downarrow$ & Flow Angular $\downarrow$ & TNAE $\downarrow$ & Trajectory Position Error $\downarrow$ \\
\midrule
DGMesh & 0.007823 & 20.42 & 15.08 & 0.00637 \\
Ours   & \textbf{0.007237} & \textbf{20.13} & \textbf{11.30} & \textbf{0.00508} \\
\bottomrule
\end{tabular}
\end{adjustbox}
\end{table*}

\noindent\textbf{Qualitative Results.}
Fig. \ref{fig:qualitative} provides qualitative comparisons against baseline methods \cite{dynamic2dgs, gags, dgmesh} on the D-NeRF \cite{dnerf} and DG-Mesh \cite{dgmesh} datasets. Standard GS-based baselines exhibit severe geometric artifacts, floating primitives, and fragmented surfaces under complex motions. While DG-Mesh \cite{dgmesh} relies on direct mesh extraction from discrete Gaussians, it causes degraded geometric details and rendering quality. D-2DGS \cite{dynamic2dgs} also misses high-frequency details, as its meshes are built upon an estimated depth model. In contrast, our method reconstructs smooth and temporally consistent deformation while preserving fine-grained geometric details.

Furthermore, Fig. \ref{fig:quantitative2} demonstrates the qualitative results of our method on the real-world Nerfies \cite{nerfies} dataset. Due to real-world data noises, including casual camera trajectories, complex backgrounds, and highly textured objects, DG-Mesh \cite{dgmesh} produces noisy surfaces, failing to capture stable geometric details as it extracts surfaces from noisy Gaussian primitives. In contrast, our method captures high-frequency details of the mesh surface with a stable deformation.
% Real-world captures introduce significant challenges, including casual camera trajectories, complex background clutter, and highly textured objects. Under these conditions, 

\subsection{Ablation Study}
\begin{figure*}[t]
\vspace{-0.2cm}
  \centering
  \includegraphics[width=\textwidth]{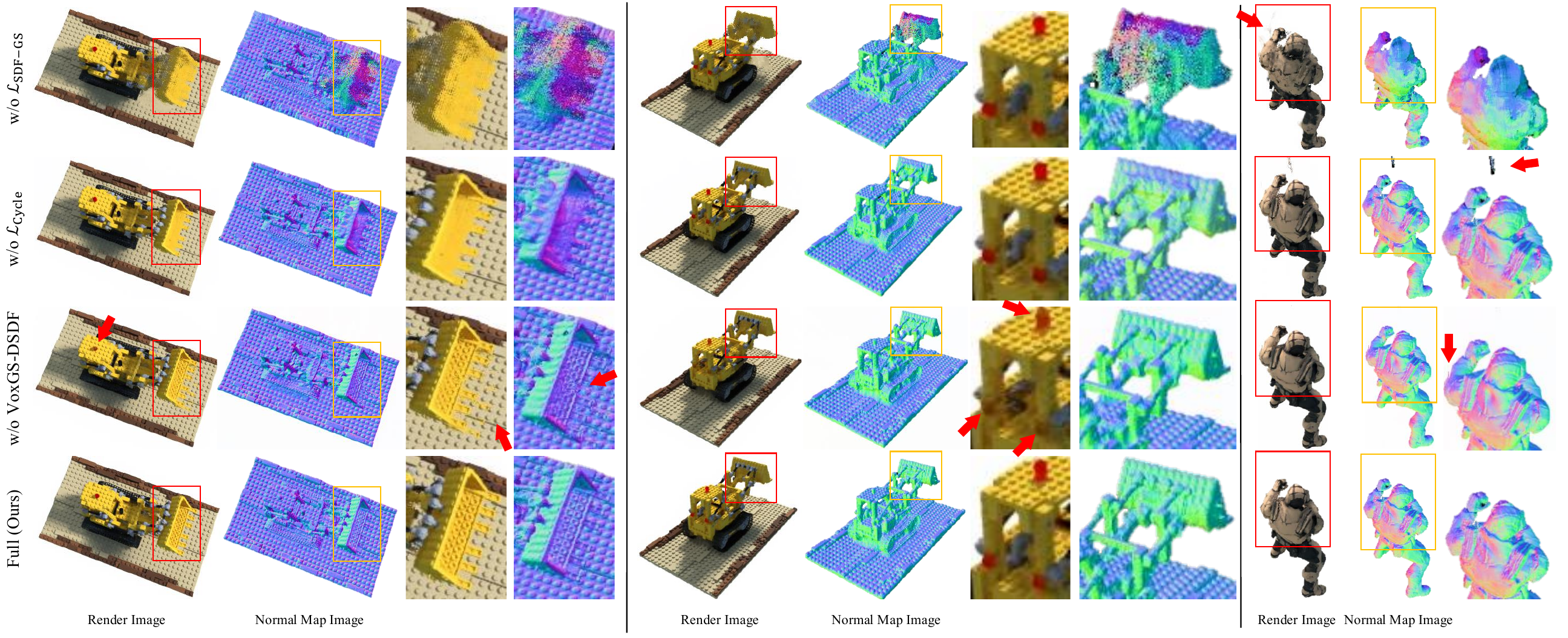}
  \caption{\textbf{Qualitative results of the ablation study.} We show the rendered image and normal map image results from the VoxGS-DSDF branch. Zoom in to check the details.}
  \label{fig:abl}
  \vspace{-0.2cm}
\end{figure*}

We perform an ablation study to analyze the individual contributions of the core components of the DySurface. Quantitative and qualitative evaluations are shown in Tab. \ref{tab:ablation} and Fig. \ref{fig:abl}, respectively.

\noindent\textbf{Effect of VoxGS-DSDF.} To validate the efficacy of our VoxGS-DSDF branch, we ablate the deformed Gaussian voxelization and ray-query matching mechanisms. In this baseline configuration, the spatial condition latent $\eta_d$ relies on the standard positional encoding of the dynamic query point $q_d$, operating without any explicit guidance. As shown in both quantitative and qualitative results, omitting this module results in a clear degradation of geometric accuracy and photometric fidelity. This confirms that embedding explicitly deformed Gaussian attributes as structural priors into the implicit function effectively improves the SDF optimization.

\noindent\textbf{Effect of SDF-GS Anchoring Loss.} The removal of the explicit anchoring loss ($w/o~\mathcal{L}_{SDF-GS}$) induces a drastic performance drop in both geometric accuracy and photometric fidelity. In the absence of this structural constraint, the implicit SDF field operates independently of the explicit voxelized Gaussians, leading to ill-posed geometric optimization. Consequently, the reconstructed surfaces exhibit severe fragmentation and floating artifacts. This highlights the necessity of our explicit-to-implicit anchoring formulation for establishing robust, watertight boundaries.

%  on SDF-GS anchoring and cycle consistency loss.
\begin{table*}[h]
\centering
\caption{\textbf{Ablation study of SDF-GS anchoring and cycle consistency loss.} The results are reported with the average of each scene.}
\label{tab:ablation}
\begin{adjustbox}{scale=0.86}
\begin{tabular}{@{}ccc|ccc@{}}
\toprule
                                & \: vIoU ($\uparrow$) \:     & \: CD ($\downarrow$) \:  & \: PSNR ($\uparrow$)  & SSIM ($\uparrow$) & LPIPS ($\downarrow$)   \:      \\ \midrule
w/o $\mathcal{L}_{SDF-GS}$ \:   &  0.3388            &  0.0248         & 24.15    & 0.9308   &  0.0420     \\
w/o $\mathcal{L}_{Cycle}$       &  0.3519            &  0.0197         & 27.50    & 0.9558   &  0.0298     \\
w/o VoxGS-DSDF                  &  0.3726            &  0.0120         & 27.22    &  0.9602  &  0.0297     \\ \midrule
Full (Ours)                     & \textbf{0.3932} & \textbf{0.0102} & \textbf{31.09} & \textbf{0.9735}  & \textbf{0.0161}  \\ \bottomrule
\end{tabular}
\end{adjustbox}
\end{table*}

\noindent\textbf{Effect of Cycle Consistency Loss.} Disabling the cycle consistency loss ($w/o$ $\mathcal{L}_{Cycle}$) yields a significant degradation in both geometric precision and photometric rendering fidelity. Notably, using the cycle consistency loss increases the vIoU from 0.3519 to 0.3932, within a corresponding increase in image-level evaluation metrics. Although the base geometry remains constrained by the VoxGS-DSDF branch, removing bidirectional regularization reduces the model's capacity to track complex deformations. As illustrated in Fig. \ref{fig:abl}, high-frequency details are significantly enhanced with the $\mathcal{L}_{Cycle}$, demonstrating that $\mathcal{L}_{Cycle}$ is an effective regularizer, ensuring spatio-temporal coherence for 3D structures and enhancing the visual quality of synthesized views.

% \subsection{Ablation Study}
% We conduct an ablation study to validate the individual contributions of our core components in the DySurface framework. The results are depicted quantitatively in Tab. \ref{tab:ablation} and qualitatively in Fig. \ref{fig:abl}.

% \subsubsection{Effect of SDF-GS Anchoring Loss.} Removing the explicit anchoring loss ($w/o~\mathcal{L}_{SDF-GS}$) results in the most catastrophic degradation across both geometric and photometric metrics. Without this constraint, the continuous implicit SDF field loses its structural dependency on the voxelized Gaussian features and suffers from ambiguous geometry. Qualitative results show that the geometries are disconnected, creating floating artifacts. This strictly proves that our explicit-to-implicit anchoring mechanism is the most critical factor in guaranteeing a watertight and stable topology.

% \subsubsection{Effect of Cycle Consistency Loss.} We perform omitting the cycle consistency loss ($w/o~\mathcal{L}_{Cycle}$), which leads to a noticeable drop in geometric precision, with vIoU decreasing from 0.3519 to 0.3917. While the primary surface topology is still roughly maintained by the SDF-GS guidance, the absence of bidirectional temporal constraints causes the network to struggle with complex, non-rigid motions. This demonstrates that $\mathcal{L}_{Cycle}$ acts as an essential regularizer for enforcing strict spatiotemporal coherence and refining high-frequency dynamic details across frames.

\vspace{-0.1cm}
\section{Conclusion}
\vspace{-0.1cm}
In this paper, we introduced DySurface, a novel hybrid framework designed to achieve high-fidelity and temporally consistent 3D surface reconstruction in dynamic scenes. By bridging the structural gap between the explicit 3D Gaussian Splatting and the geometric fidelity of implicit Signed Distance Functions (SDF), our method robustly extracts highly detailed deformable surfaces. Our core VoxGS-DSDF architecture explicitly anchors the continuous SDF field using dynamically deformed sparse voxels, robustly resolving geometric ambiguities during deformation. Comprehensive evaluations across synthetic and real-world datasets support that DySurface achieves significant performance in both geometric accuracy and photometric quality, successfully producing watertight meshes rather than fragmented proxies. We believe our approach provides a step forward for reliable 4D content creation. Our future work explores accelerating the optimization process and extending the framework to accommodate severe topological changes, such as tearing and merging.

\section{Acknowledgement}
This work was in part supported by NST grant (CRC 21015, MSIT), IITP grant (RS-2023-00228996, RS-2024-00459749, RS-2025-25443318, RS-2025-25441313, MSIT) and InnoCORE program (N10260110, MSIT).

%%%%%%%%%%%%%%%%%%%%%%%%%%%%%%%%%%%%%%%%%%%%%%%%%%%%%%%%%%%%
% \newpage

\bibliographystyle{plain}
\bibliography{main}

%%%%%%%%%%%%%%%%%%%%%%%%%%%%%%%%%%%%%%%%%%%%%%%%%%%%%%%%%%%%

\newpage
\appendix

% \section{Technical appendices and supplementary material}
% Technical appendices with additional results, figures, graphs, and proofs may be submitted with the paper submission before the full submission deadline (see above). You can upload a ZIP file for videos or code, but do not upload a separate PDF file for the appendix. There is no page limit for the technical appendices. 

% Note: Think of the appendix as ``optional reading'' for reviewers. The paper must be able to stand alone without the appendix; for example, adding critical experiments that support the main claims to an appendix is inappropriate. 
\section{Supplementary Material}
Here, we present additional experimental results with DySurface that are not included in the main paper. We provide detailed descriptions of DySurface and several additional experimental results.

\section{Additional Implementation Details}
\subsection{Network Architecture}
\subsubsection{Geometry Representation (SDF).}
To represent the continuous implicit surface, we utilize a multi-resolution hash grid architecture. The spatial coordinates are encoded using a 16-level progressive band hash grid with a base resolution of 32, 2 feature dimensions per level, and a maximum hash map size of $2^{21}$. The Geometry MLP consists of 1 hidden layer with 128 neurons and employs Softplus activation along with weight normalization. To ensure stable early training, we apply geometric initialization to approximate an initial sphere with a radius of 0.8. The network outputs a 128-dimensional geometric feature alongside the SDF value. The color network predicts RGB values using the 128-dimensional geometric feature, time embeddings, and viewing directions encoded via Spherical Harmonics (degree of 4). The color network is composed of 2 hidden layers with 128 neurons each, utilizing ReLU activations for intermediate layers and a final Sigmoid activation to bound the RGB output.

\subsubsection{Sparse Voxel \& Dynamic Deformation.}
\begin{wraptable}{r}{0.6\textwidth} % 'l'은 표를 왼쪽에 배치, '0.5\textwidth'는 표가 차지할 가로 너비 비율
  \centering
    \caption{Loss weight configurations across the three progressive training stages of DySurface. Symbols correspond to those defined in the main paper. RGB rendering losses ($\mathcal{L}_{rgb}$, $\mathcal{L}_{rgb}^{sdf}$, 
$\mathcal{L}_{rgb}^{mesh}$) use a weight of 1.0 by default in all stages.}
    \begin{adjustbox}{scale=0.92}
    \label{tab:loss_weights}
    \renewcommand{\arraystretch}{1}
    \begin{tabular}{llcc}
    \toprule
    \textbf{Stage} & \textbf{Symbol} & \textbf{Loss Term} & \textbf{Value} \\
    \midrule
    \multirow{3}{*}{\shortstack[l]{GS Branch}}
     & $\lambda_{m}$        & $\mathcal{L}_{mask}$       & 0.3  \\
     & $\lambda^{gs}_{reg}$ & $\mathcal{L}_{reg}^{gs}$   & 1.0  \\
    \midrule
    \multirow{6}{*}{\shortstack[l]{VoxGS-DSDF \\ Branch}}
     & $\lambda_{c}$     & $\mathcal{L}_{cycle}$      & 1.0 \\
     & $\lambda_{SG}$    & $\mathcal{L}_{SDF\text{-}GS}$ & 1.0 \\
     & $\lambda_{dist}$  & $\mathcal{L}_{distill}$    & 1.0  \\
     & $\lambda_{e}$     & $\mathcal{L}_{eik}$        & 0.1  \\
     & $\lambda_{s}$     & $\mathcal{L}_{smooth}$     & 1.0  \\
    \midrule
    \multirow{4}{*}{\shortstack[l]{Mesh \\ Refinement}}
     & $\lambda^{mesh}_{m}$  & $\mathcal{L}_{mask}^{mesh}$ & 0.5  \\
     & $\lambda_{lap}$       & $\mathcal{L}_{lap}$         & 1.0 \\
     & $\lambda^{mesh}_{reg}$& $\mathcal{L}_{reg}^{mesh}$  & 0.1 \\
    \bottomrule
    \end{tabular}
    \end{adjustbox}
\end{wraptable}

We anchor the Gaussian features into a sparse volumetric representation. The canonical space is bounded by a sparse voxel grid configured with a maximum resolution of $256^3$ and a voxel size of 0.002. Each valid voxel stores a 6-dimensional latent feature vector. We apply an occupancy threshold of 0.001 to aggressively prune empty grid cells and maintain computational efficiency. Instead of heuristic functional bases, the non-rigid dynamics are modeled through a dedicated Delta network. The continuous deformation from the canonical to the dynamic space is parameterized by mapping the sparse voxel features through this Delta network, ensuring structurally coherent motion tracking. 

For volume rendering, we dynamically sample up to 1024 points per ray, which includes 256 fine samples to capture detailed topological structures accurately. Its dynamic field motion is learned using the same network structure as the Gaussian Splatting branch.

\subsection{Hyperparameters and Training Details}

Our framework is trained progressively in three stages using the AdamW optimizer. Each prior stage is frozen during the next stage optimization. The Gaussian Splatting branch is trained for 30,000 steps, followed by 30,000 steps for the VoxGS-DSDF branch, and a final 1,000 steps for the Mesh Refinement branch. The base learning rate is set to 0.01 for the texture network, while the geometry and deformation modules use 0.001. A sequential learning rate scheduler is applied, multiplying the learning rate by 0.1 at 20,000 steps, with additional decay at later milestones.

Due to space limitations in the main paper, we provide the complete loss weight configurations across all three training stages in Table~\ref{tab:loss_weights}. The reported values are fixed across all evaluated scenes, and the majority of weights are simply set to 1.0; per-scene configurations, when adjusted in specific scenes, will be released in our codebase.

In Eq. (8), $\epsilon$ serves as a spatial distance threshold that defines the boundary proximity for anchoring the implicit zero-level set to the explicit canonical Gaussian. We set $\epsilon$ relative to the overall spatial scale of the scene. For the D-NeRF benchmark, we set $\epsilon=1 \times 10^{-4}$, which corresponds to approximately 0.008 of the object's bounding radius. Setting $\epsilon$ too large relaxes the geometric anchoring constraint, allowing the implicit SDF field to drift away from explicit Gaussians and causing loss of sharp surface details. Conversely, setting $\epsilon$ too small overly constrains the continuous SDF field to discrete Gaussian centers, which can introduce local surface noise. Our relative scaling choice balances smooth zero-level set continuity with sharp explicit boundary alignment.

\subsection{Details of Temporal Smoothness Term}
\label{sec:temporal_smoothness}

To encourage temporally smooth deformation in the VoxGS-DSDF branch, we regularize the predicted backward deformation offsets of ray-sampled query points. Let $\Delta \mathbf{q}_i(t)$ denote the predicted deformation offset of the $i$-th query point at time $t$. Given its neighboring timestamps $t_{\mathrm{prev}}$ and $t_{\mathrm{next}}$, we randomly sample intermediate timestamps as
\begin{equation}
\tau_{\mathrm{prev}} = t+\alpha(t_{\mathrm{prev}}-t), \qquad \alpha \sim \mathcal{U}(0,1),
\end{equation}
and
\begin{equation}
\tau_{\mathrm{next}} = t+\beta(t_{\mathrm{next}}-t), \qquad \beta \sim \mathcal{U}(0,1).
\end{equation}

The temporal smoothness loss is defined as
\begin{equation}
\mathcal{L}_{\mathrm{smooth}} = \mathcal{L}_{\mathrm{smooth}}^{\mathrm{prev}} + \mathcal{L}_{\mathrm{smooth}}^{\mathrm{next}},
\end{equation}
where
\begin{equation}
\mathcal{L}_{\mathrm{smooth}}^{\mathrm{prev}} = \frac{1}{N} \sum_{i=1}^{N} \left\| \Delta \mathbf{q}_i(t) - \overline{\Delta \mathbf{q}_i(\tau_{\mathrm{prev}})} \right\|_2^2,
\end{equation}
and
\begin{equation}
\mathcal{L}_{\mathrm{smooth}}^{\mathrm{next}} = \frac{1}{N} \sum_{i=1}^{N} \left\| \Delta \mathbf{q}_i(t) - \overline{\Delta \mathbf{q}_i(\tau_{\mathrm{next}})} \right\|_2^2.
\end{equation}

Here, $N$ denotes the number of ray-sampled query points, and the overline denotes a gradient-detach operation. The deformation offsets predicted at the intermediate timestamps are treated as detached temporal targets, encouraging the backward deformation field to vary smoothly over time.

% \subsection{Hyperparameters}
% Our framework is trained for 30,000 steps for the Gaussian Splatting branch, another 30,000 steps for the VoxGS-DSDF branch, and 1,000 steps for the Mesh Refining branch using the AdamW optimizer. The base learning rate for the texture network is set to 0.01, while the geometry and deformation modules use a learning rate of 0.001. A sequential learning rate scheduler is applied, multiplying the learning rate by 0.1 at 20,000 steps, with additional decay applied at later milestones.

% Due to space limitations in the main paper, we detail the exact hyperparameter configurations utilized across the three progressive training stages of DySurface. The total loss is formulated as a weighted sum of multiple objectives. Key hyperparameter weights include $\lambda^{sdf}_{rgb} = 1.0$, $\lambda^{sdf}_{mask} = 0.3$, and $\lambda_{eikonal} = 0.1$. For dynamic consistency and structural regularization, we assign $\lambda_{cycle} = 10.0$, $\lambda_{SDF-GS} = 1.0$, and $\lambda_{lap} = 60$. Furthermore, to ensure high-fidelity surface extraction, the mesh regularization terms are set with $\lambda_{mesh\_rendering} = 1.0$ and $\lambda_{mesh\_reg} = 10.0$. The precise configurations across scenes will be fully documented in our released codebase.

% While the aforementioned values represent our default configuration, minor scene-specific empirical adjustments are applied to accommodate varying degrees of motion complexity, scale, and topological changes. 

\section{Additional Experiments}
\subsection{Evaluation Across Time Steps} 
Unlike prior works such as GaGS \cite{sgags}, DG-Mesh \cite{sdgmesh}, 4DGS \cite{s4dgs}, and FDNeRF \cite{sfdnerf}, which only quantitatively measure novel view synthesis photometric quality, we are the first to explicitly measure and report the 3D geometry accuracy on the D-NeRF \cite{sdnerf} dataset. 

We present additional quantitative results to provide a fully transparent and granular breakdown of our framework's performance. Tab. \ref{tab:supplement_grouped_by_time} details the geometric evaluation scores (vIoU and Chamfer Distance) grouped by distinct continuous time steps ($t \in \{0.0, 0.5, 1.0\}$). Within each time step, we provide the metric for individual scenes alongside their cross-scene averages to demonstrate temporal stability. Furthermore, the overall photometric metrics (PSNR, SSIM, and LPIPS) are reported at the bottom of the table. This discrete time step breakdown confirms that our method maintains robust reconstruction fidelity and structural coherence not only near the canonical space but also at distant temporal states.

\begin{table}[htbp]
\centering
\caption{\textbf{Comprehensive Detailed Quantitative Evaluation by Time Step.} We report the detailed metrics grouped by each specific time step ($t \in \{0.0, 0.5, 1.0\}$). Methods in parentheses indicate the mesh extraction method. HW stands for HellWarrior. }
\label{tab:supplement_grouped_by_time}
\resizebox{\textwidth}{!}{
\begin{tabular}{l | cccc|ccc}
\toprule
\textbf{Metric} & TiNeuVox & \shortstack{4DGS +\\(3DGS-to-PC)} & \shortstack{GaGS +\\(GS2Mesh)} & \shortstack{GaGS +\\(3DGS-to-PC)} & DG-Mesh &D-2DGS   & \textbf{Ours} \\
\midrule
\multicolumn{7}{c}{\textbf{\textit{$t=0.0$}}} \\
\midrule
vIoU (Lego) $\uparrow$ & 0.1300 & 0.2252 & 0.3286 & 0.2446 & 0.2614 & 0.2196 & 0.3012 \\
vIoU (T-Rex) $\uparrow$ & 0.5545 & 0.2764 & 0.0783 & 0.2878 & 0.5693 & 0.4949 & 0.5532 \\
vIoU (HW) $\uparrow$ & 0.1991 & 0.2454 & 0.1990 & 0.2347 & 0.3037  & 0.3047 & 0.2906 \\
\cmidrule{2-8}
\textbf{vIoU (Avg)} $\uparrow$ & 0.2945 & 0.2490 & 0.2020 & 0.2557 & \underline{0.3781} & 0.3397 & \textbf{0.3817} \\
\cmidrule{1-8}
CD (Lego) $\downarrow$ & 0.0180 & 0.0188 & 0.0170 & 0.0157 & 0.0242 & 0.0128 & 0.0132 \\
CD (T-Rex) $\downarrow$ & 0.0073 & 0.0700 & 0.0211 & 0.0068 & 0.0046 & 0.0056 & 0.0051 \\
CD (HW) $\downarrow$ & 0.0272 & 0.0232 & 0.0434 & 0.0230 & 0.0201  & 0.0135 & 0.0129 \\
\cmidrule{2-8}
\textbf{CD (Avg)} $\downarrow$ & 0.0175 & 0.0373 & 0.0272 & 0.0150 & 0.0163 & \underline{0.0106} & \textbf{0.0104} \\
\midrule
\multicolumn{8}{c}{\textbf{\textit{$t=0.5$}}} \\
\midrule
vIoU (Lego) $\uparrow$ & 0.1080 & 0.2210 & 0.2772 & 0.2289 & 0.2701 & 0.2200 & 0.3015 \\
vIoU (T-Rex) $\uparrow$ & 0.1634 & 0.2542 & 0.0517 & 0.2198 & 0.5779 & 0.5029 & 0.6285 \\
vIoU (HW) $\uparrow$ & 0.3122 & 0.2168 & 0.1566 & 0.2586 & 0.3164   & 0.3162  & 0.3101 \\
\cmidrule{2-8}
\textbf{vIoU (Avg)} $\uparrow$ & 0.1945 & 0.2307 & 0.1618 & 0.2358 & \underline{0.3881} & 0.3464  & \textbf{0.4134} \\
\cmidrule{1-8}
CD (Lego) $\downarrow$ & 0.0174 & 0.0187 & 0.0171 & 0.0164 & 0.0242 & 0.0139 & 0.0139 \\
CD (T-Rex) $\downarrow$ & 0.0068 & 0.0680 & 0.0275 & 0.0071 & 0.0042 & 0.0053 & 0.0048 \\
CD (HW) $\downarrow$ & 0.0226 & 0.0216 & 0.0472 & 0.0185 & 0.0189  & 0.0121  & 0.0115 \\
\cmidrule{2-8}
\textbf{CD (Avg)} $\downarrow$ & 0.0156 & 0.0361 & 0.0306 & 0.0140 & 0.0158 & \underline{0.0104} & \textbf{0.0101} \\
\midrule
\multicolumn{8}{c}{\textbf{\textit{$t=1.0$}}} \\
\midrule
vIoU (Lego) $\uparrow$ & 0.0860 & 0.2001 & 0.2699 & 0.2191 & 0.2649 & 0.2019 & 0.2938 \\
vIoU (T-Rex) $\uparrow$ & 0.5518 & 0.2799 & 0.0620 & 0.2513 & 0.5735 & 0.5014 & 0.5532 \\
vIoU (HW) $\uparrow$ & 0.2292 & 0.2266 & 0.1934 & 0.2347 & 0.3002  & 0.3029  & 0.2841 \\
\cmidrule{2-8}
\textbf{vIoU (Avg)} $\uparrow$ & 0.2890 & 0.2355 & 0.1751 & 0.2350 & \textbf{0.3795} &  0.3354 & \underline{0.3770} \\
\cmidrule{1-8}
CD (Lego) $\downarrow$ & 0.0168 & 0.0166 & 0.0160 & 0.0145 & 0.0225 & 0.0139 & 0.0129 \\
CD (T-Rex) $\downarrow$ & 0.0072 & 0.0750 & 0.0492 & 0.0069 & 0.0045 & 0.0055 & 0.0052 \\
CD (HW) $\downarrow$ & 0.0260 & 0.0215 & 0.0403 & 0.0230 & 0.0206 & 0.0139 & 0.0128 \\
\cmidrule{2-8}
\textbf{CD (Avg)} $\downarrow$ & 0.0167 & 0.0377 & 0.0352 & 0.0148 & 0.0158  & \underline{0.0111} & \textbf{0.0103} \\
\midrule
\multicolumn{8}{c}{\textbf{\textit{Overall Metrics}}} \\
\midrule
\textbf{PSNR (Avg)} $\uparrow$ & 28.71 & 28.97 & \multicolumn{2}{c}{\textbf{32.24}} & 25.23 & 25.82 & \underline{31.09} \\
\cmidrule{1-8}
\textbf{SSIM (Avg)} $\uparrow$ & 0.9554 & 0.9678 & \multicolumn{2}{c}{\textbf{0.9754}} & 0.9187 &  0.9374  & \underline{0.9735} \\
\cmidrule{1-8}
\textbf{LPIPS (Avg)} $\downarrow$ & 0.0542 & 0.0261 & \multicolumn{2}{c}{\underline{0.0182}} & 0.1027 & 0.0681  & \textbf{0.0161} \\
\bottomrule
\end{tabular}
}
\end{table}

\begin{table*}[htbp] % 하나의 큰 플로팅 환경으로 묶음
\centering

% ------------------- [ 1. 테이블 부분 ] -------------------
\caption{\textbf{Quantitative comparison on the D-NeRF~\cite{sdnerf} dataset.} We compare our method against state-of-the-art baselines. The best results are highlighted in \textbf{bold}.}
\label{tab:dnerf_comparison}
\resizebox{\textwidth}{!}{
\begin{tabular}{@{}l ccc ccc ccc ccc@{}}
\toprule
& \multicolumn{3}{c}{Hell Warrior} & \multicolumn{3}{c}{Mutant} & \multicolumn{3}{c}{Hook} & \multicolumn{3}{c}{Bouncing Balls} \\
\cmidrule(lr){2-4} \cmidrule(lr){5-7} \cmidrule(lr){8-10} \cmidrule(lr){11-13}
Method & PSNR$\uparrow$ & SSIM$\uparrow$ & LPIPS$\downarrow$ & PSNR$\uparrow$ & SSIM$\uparrow$ & LPIPS$\downarrow$ & PSNR$\uparrow$ & SSIM$\uparrow$ & LPIPS$\downarrow$ & PSNR$\uparrow$ & SSIM$\uparrow$ & LPIPS$\downarrow$ \\
\midrule
3D-GS           & 15.39 & 0.8776 & 0.1300      & 21.75 & 0.9359 & 0.0575            & 18.69 & 0.8733 & 0.1144        & 22.56 & 0.9485 & 0.0647 \\
D-NeRF          & 25.02 & 0.9506 & 0.0691      & 31.29 & 0.9739 & 0.0268            & 29.26 & 0.9650 & 0.1174        & 38.93 & 0.9900 & 0.1031 \\
TiNeuVox        & 28.20 & 0.9661 & 0.0631      & 33.90 & 0.9771 & 0.0301            & 31.79 & 0.9718 & 0.0436        & 40.85 & 0.9913 & 0.0401 \\
NDVG            & 26.49 & 0.9600 & 0.0670      & 34.41 & 0.9801 & 0.0270            & 30.00 & 0.9626 & 0.0463        & 37.52 & 0.9874 & 0.0751 \\
FDNeRF          & 27.71 & 0.9665 & 0.0508      & 34.97 & 0.9810 & 0.0312            & 32.29 & 0.9756 & 0.0388        & 40.02 & 0.9912 & 0.0395 \\
4D-GS           & 28.12 & 0.9730 & 0.0276      & 38.34 & 0.9936 & 0.0062            & 33.16 & 0.9810 & 0.0168        & 40.74 & 0.9941 & 0.0105 \\
GaGS            & \textbf{32.27} & \textbf{0.9835} & \textbf{0.0164}     & \textbf{41.43} & \textbf{0.9969} & \textbf{0.0029}         & \textbf{36.96} & \textbf{0.9916} & \textbf{0.0076}           & \textbf{43.59} & \textbf{0.9960} & \textbf{0.0061} \\  \midrule
DG-Mesh         & 25.46 & 0.9590 & 0.0842      & 30.40 & 0.9680 & 0.0550            & 27.88 & 0.9540 & 0.0740       & 29.14 & 0.9690 & 0.0990 \\
D-2DGS         & 25.50 & 0.9583 & 0.1121      & 28.12 & 0.9600 & 0.0423            & 27.80 & 0.9620 & 0.0419        & 27.78 & 0.9690 & 0.0730 \\
Ours            & 31.14 & \textbf{0.9835} & 0.0169  & 32.25 & 0.9833  &  0.0099      & 32.00   & 0.9822  & 0.0148   & 35.99  &  0.9856 & 0.0136  \\
\midrule
& \multicolumn{3}{c}{Lego} & \multicolumn{3}{c}{T-Rex} & \multicolumn{3}{c}{Stand Up} & \multicolumn{3}{c}{Jumping Jacks} \\
\cmidrule(lr){2-4} \cmidrule(lr){5-7} \cmidrule(lr){8-10} \cmidrule(lr){11-13}
Method & PSNR$\uparrow$ & SSIM$\uparrow$ & LPIPS$\downarrow$ & PSNR$\uparrow$ & SSIM$\uparrow$ & LPIPS$\downarrow$ & PSNR$\uparrow$ & SSIM$\uparrow$ & LPIPS$\downarrow$ & PSNR$\uparrow$ & SSIM$\uparrow$ & LPIPS$\downarrow$ \\
\midrule
3D-GS           & 23.10 & 0.9329 & 0.0567        & 25.75 & 0.9567 & 0.0474       & 19.38 & 0.9200 & 0.0909       & 20.72 & 0.9227 & 0.0980 \\
D-NeRF          & 21.64 & 0.8394 & 0.1654        & 31.75 & 0.9767 & 0.0396       & 32.80 & 0.9818 & 0.0215       & 32.80 & 0.9810 & 0.0373 \\
TiNeuVox        & 25.17 & 0.9217 & 0.0689        & 32.77 & 0.9783 & 0.0307       & 36.20 & 0.9859 & 0.0199       & 34.74 & 0.9823 & 0.0328 \\
NDVG            & 25.04 & 0.9395 & 0.0534        & 32.62 & 0.9781 & 0.0330       & 33.22 & 0.9793 & 0.0302       & 31.25 & 0.9737 & 0.0398 \\
FDNeRF          & 25.27 & 0.9390 & 0.0460        & 30.71 & 0.9731 & 0.0368       & 36.91 & 0.9878 & 0.0188       & 33.55 & 0.9812 & 0.0329 \\
4D-GS           & 25.40 & 0.9434 & 0.0377        & 33.39 & 0.9869 & 0.0130       & 38.26 & 0.9923 & 0.0071       & 35.67 & 0.9882 & 0.0159 \\
GaGS            & \textbf{25.44} & \textbf{0.9474} & 0.0329            & \textbf{39.03} & \textbf{0.9952} & \textbf{0.0052}                & \textbf{42.21} & \textbf{0.9966} & \textbf{0.0028}                & \textbf{37.96} & \textbf{0.9928} & \textbf{0.0088} \\ \midrule
DG-Mesh         & 21.29 & 0.8380 & 0.1590        & 28.95 & 0.9590 & 0.0650       & 30.21 & 0.9740 & 0.0510       & 31.77 & 0.9770 & 0.0450 \\
D-2DGS         & 23.29 & 0.8872 & 0.1124        & 28.67 & 0.9670 & 0.0434       & 29.51 & 0.9768 & 0.0280       & 29.29 & 0.9741  & 0.0322 \\
Ours            & 25.40 & 0.9439 & \textbf{0.0245}                & 36.75 & 0.9932 & 0.0069       & 36.08  & 0.9903 & 0.0089                     &  32.35 &   0.9822  & 0.0171   \\
\bottomrule
\end{tabular}
} % resizebox 끝

\vspace{1.0cm} % 표와 그림 사이의 수직 간격 조절

% ------------------- [ 2. 그림 부분 ] -------------------
% 별도의 figure 환경을 열지 않고 현재 table* 환경 안에서 이미지를 삽입합니다.
\includegraphics[width=\textwidth]{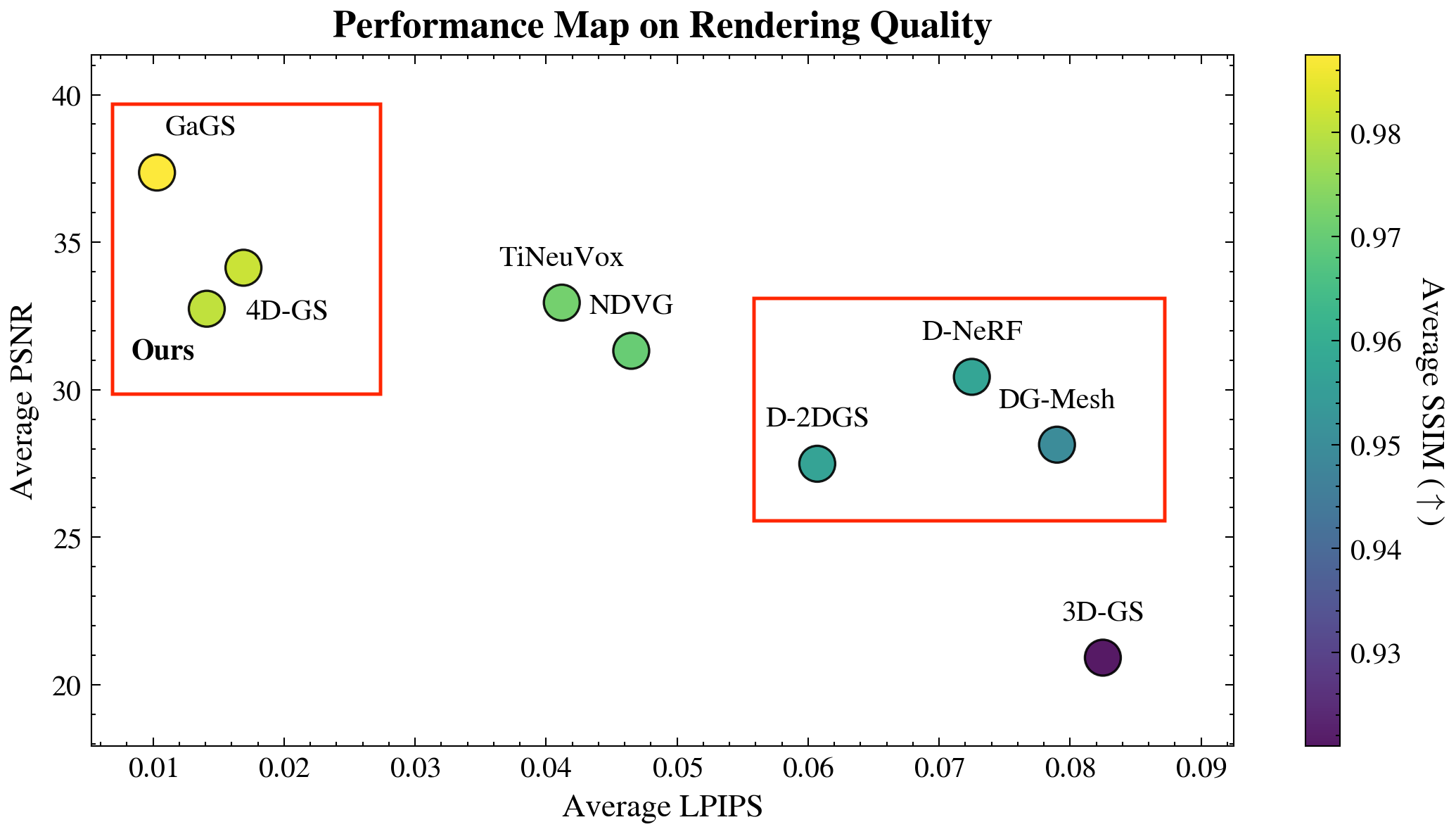}

\captionof{figure}{\textbf{Average performance map on the D-NeRF dataset.} This scatter plot visualizes the quantitative results averaged over all eight scenes. The x-axis and y-axis represent the average LPIPS and PSNR, respectively, while the marker color indicates the average SSIM score. Our method is highlighted in bold.}
\label{sup:fig:plot}
\end{table*}

\subsection{Results on Remaining Scenes}
\begin{figure*}[htbp]
  \centering
  \includegraphics[width=\textwidth]{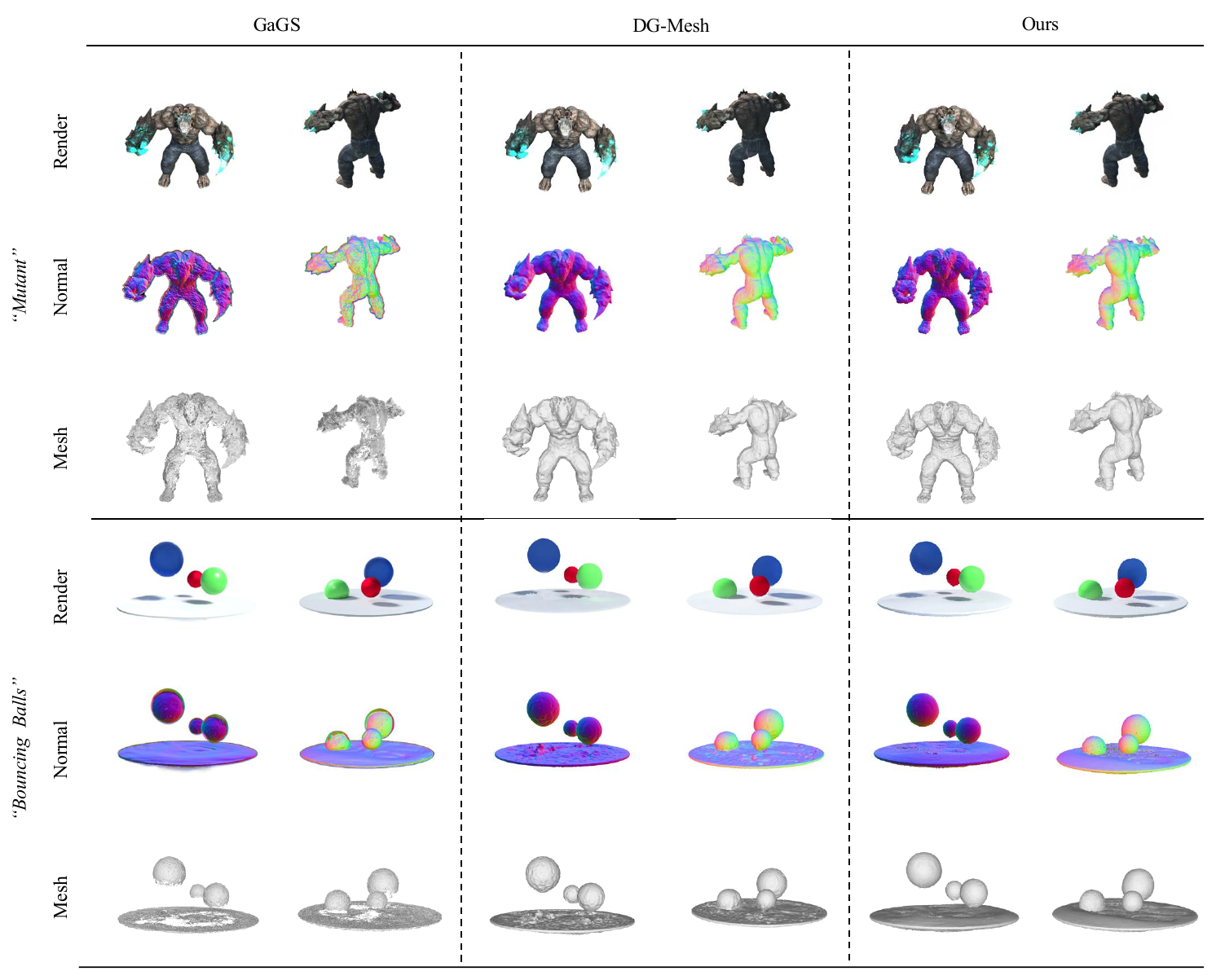}
  \caption{\textbf{Additional qualitative comparison results on diverse D-NeRF scenes.} We show the rendered image, normal map image, and mesh surface image results of GaGS \cite{sgags}, DG-Mesh \cite{sdgmesh}, and DySurface (Ours). Please zoom to check the details.}
  \label{sup:fig:remaining_scenes1}
\end{figure*}

In Fig. \ref{sup:fig:remaining_scenes1}, Fig. \ref{sup:fig:remaining_scenes2}, and Fig. \ref{sup:fig:remaining_scenes3}, we provide visual results on the remaining scenes of the D-NeRF \cite{sdnerf} dataset. The qualitative results demonstrate that DySurface consistently captures continuous deformation, high-fidelity rendering results, and preserves sharp structural details. 

We report the quantitative photometric metrics on the remaining dynamic scenes in Tab. \ref{tab:dnerf_comparison} and Fig. \ref{sup:fig:plot}. We compare against representative baselines, including 3D-GS~\cite{s3dgs}, D-NeRF~\cite{sdnerf}, TiNeuVox~\cite{stineuvox}, NDVG~\cite{ndvg}, FDNeRF~\cite{sfdnerf}, 4D-GS~\cite{s4dgs}, DG-Mesh~\cite{sdgmesh}, D-2DGS \cite{dynamic2dgs}, and GaGS~\cite{sgags}. As our framework is fundamentally designed to extract geometrically accurate and topologically coherent surfaces, it enforces spatial regularization of the continuous SDF field. Consequently, while purely NVS-dedicated methods may yield marginally higher photometric performance by freely scattering unconstrained Gaussians to overfit the training views, they fail to recover meaningful geometry. In contrast, DySurface maintains highly competitive photometric fidelity while providing state-of-the-art geometric reconstruction, achieving an optimal balance between visual rendering quality and structural 3D geometry. Fig. \ref{sup:fig:plot} also shows that our method achieves a competitive rendering score against NVS-optimized methods, compared to the prior surface reconstruction methods \cite{dynamic2dgs, dgmesh}.

\subsection{Results on Real-world Dataset}
We additionally test our method on real-world datasets to validate practical applicability. In addition to our quantitative evaluations on the real-world Nerfies [32] dataset (Table 2 and Figure 5), we conducted additional experiments on the real-world DyCheck \cite{sdycheck} benchmark across 7 different scenes \textit{(haru-sit, backpack, teddybear, pillow, creeper, mochi-high-five, sriracha-tree)}. As presented in Tab. \ref{tab:dycheck}, DySurface outperforms baseline methods in photometric quality under unconstrained camera trajectories. 

 % Tab. \ref{} demonstrates DySurface’s real-world generalization.

\begin{table}[t]
\centering
\caption{\textbf{Real-world dataset experiment on DyCheck \cite{sdycheck}.}}
\label{tab:dycheck}
\begin{tabular}{lccc}
\toprule
Method & PSNR $\uparrow$ & SSIM $\uparrow$ & LPIPS $\downarrow$ \\
\midrule
D-2DGS  & 21.91 & 0.8769 & 0.1546 \\
DG-Mesh & 22.06 & 0.8705 & 0.1858 \\
Ours    & \textbf{24.45} & \textbf{0.8871} & \textbf{0.1309} \\
\bottomrule
\end{tabular}
\end{table}

\subsection{Results on Comprehensive Frame Test}
\label{sec:comprehensive_frame}

The D-NeRF [36] benchmark is a synthetic dataset generated from dynamic 3D assets. However, the official release provides only rendered image sequences and does not include ground-truth dynamic meshes for geometric evaluation. To establish an accurate geometry benchmark consistent with the motion trajectories used in D-NeRF, we curated the original 3D source assets for \textit{Lego}, \textit{T-Rex}, and \textit{HellWarrior}, and matched their corresponding keyframe animation tracks. Using Blender, we exported the ground-truth deformed meshes over 60 consecutive timestamps covering the full motion interval, $t \in [0,1]$.

\begin{figure*}[h]
    \centering
    \includegraphics[width=\textwidth]{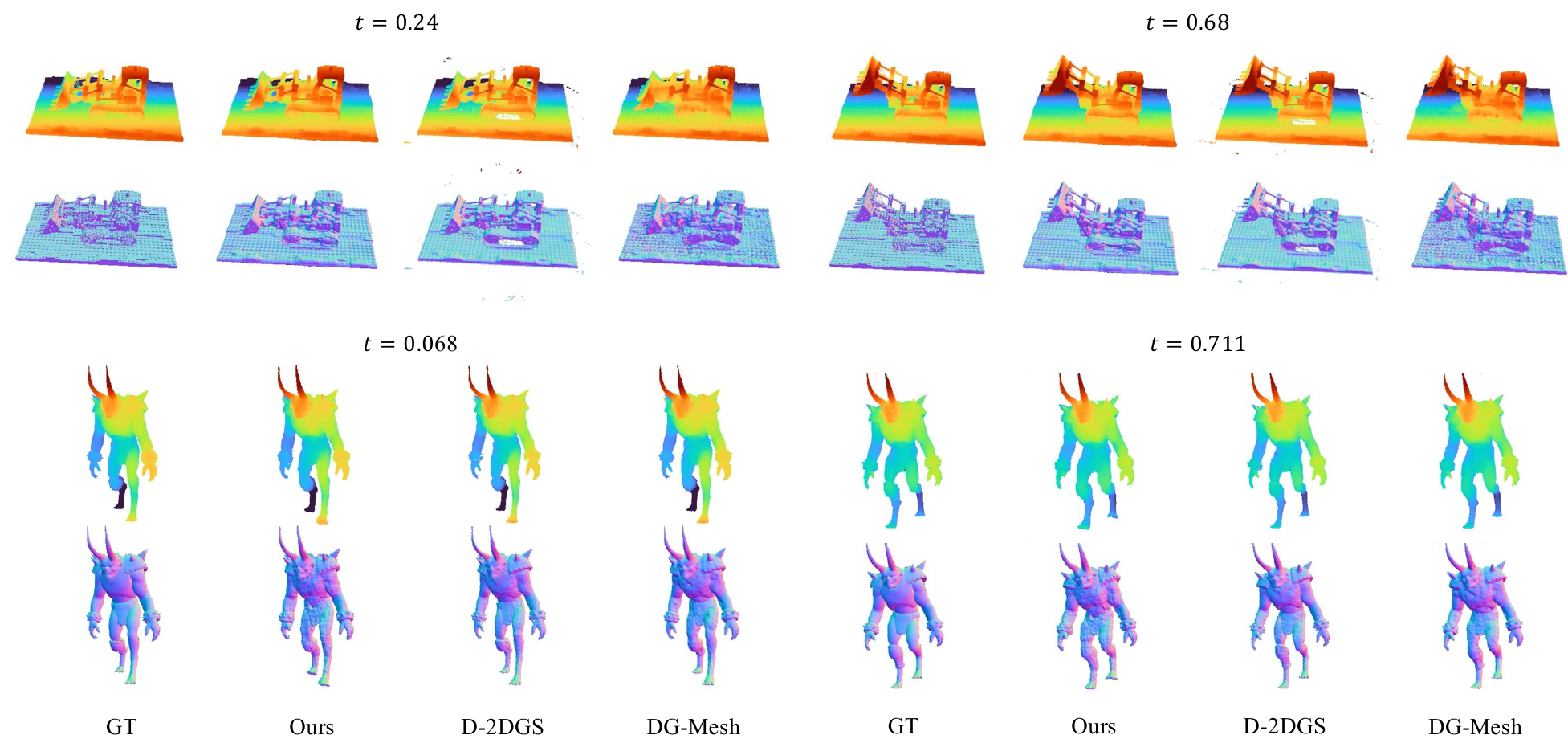}
    \caption{\textbf{Qualitative results of the dense frame test.} We visualize the depth and normal-rendered image at randomly selected times, including ground truth.}
    \label{fig:dense}
\end{figure*}

We further conduct a dense-frame geometry evaluation on all available frames of \textit{Lego}, \textit{T-Rex}, and \textit{Hell Warrior}. We report volumetric IoU (vIoU) and Chamfer Distance (CD) for 3D geometric accuracy, together with Depth Mean Absolute Error (Depth MAE) and depth accuracy within 5\% of the object scale (D-Acc@5\%) under the test view cameras. We additionally provide qualitative depth and normal renderings to visualize surface accuracy throughout the sequences. As shown in Tab. \ref{tab:dense_geometry}, our method achieves the best average performance across all four metrics, demonstrating accurate geometry reconstruction over dense temporal sequences. Fig. \ref{fig:dense} shows the qualitative result of depth and normal renderings. 

\begin{table}[t]
\centering
\caption{\textbf{Dense geometry evaluation over 60 consecutive timestamps.} vIoU and CD evaluate 3D geometric accuracy, while Depth MAE and D-Acc@5\% evaluate rendered depth accuracy.}
\label{tab:dense_geometry}
\begin{adjustbox}{scale=0.99}
\begin{tabular}{llcccc}
\toprule
Scene & Method & \:  vIoU $\uparrow$ \: & \: CD $\downarrow$ \: & DepthMAE $\downarrow$ & D-Acc@5\% $\uparrow$ \\
\midrule
\multirow{3}{*}{Lego}
& DG-Mesh & 0.2173 & 0.0731 & 0.0470 & 93.09 \\
& D-2DGS  & 0.1643 & \textbf{0.0439} &  0.0393 & \textbf{95.74} \\
& Ours    & \textbf{0.2263} & \textbf{0.0439} & \textbf{0.0384} & 95.69 \\
\midrule
\multirow{3}{*}{T-Rex}
& DG-Mesh & 0.5549 & \textbf{0.0538} & \textbf{0.0364} & 94.34 \\
& D-2DGS  & 0.4679 & 0.0565 &  0.0387 & 94.31 \\
& Ours    & \textbf{0.5608} & 0.0580 & 0.0399 & \textbf{94.45} \\
\midrule
\multirow{3}{*}{HellWarrior}
& DG-Mesh & 0.2971 & 0.0298 & \textbf{0.0225} & \textbf{97.92} \\
& D-2DGS  & \textbf{0.2983} & 0.0272 & 0.0255 & 96.95 \\
& Ours    & 0.2959 & \textbf{0.0240} & 0.0250 & 97.59 \\
\midrule
\multirow{3}{*}{Average}
& DG-Mesh & 0.3564 & 0.0522 & 0.0350 & 95.11 \\
& D-2DGS  & 0.3102 & 0.0425 & 0.0345 & 95.66 \\
& Ours    & \textbf{0.3610} & \textbf{0.0420} & \textbf{0.0344} & \textbf{95.91}\\
\bottomrule
\end{tabular}
\end{adjustbox}
\end{table}

\subsection{Effect of Gaussian Representation: 3DGS vs. 2DGS}
\label{sec:3dgs_vs_2dgs}
To investigate the effect of the Gaussian representation, we replace the 3DGS primitives in our framework with 2DGS [17], while keeping the remaining pipeline unchanged. As shown in Table~\ref{tab:3dgs_vs_2dgs}, the 3DGS-based configuration achieves better average performance in both geometric reconstruction and rendering quality. In particular, it substantially improves Chamfer Distance (CD), while maintaining comparable or better photometric quality across the evaluated scenes.

Although 2DGS provides an explicit surface-oriented representation, its primitives are constrained to planar surfels.
Under continuous non-rigid deformation, we observed optimization instability and occasional primitive flipping, which can degrade geometric reconstruction.
In contrast, 3DGS provides full 3D covariance and greater flexibility for modeling complex deformations.
Since surface regularization is already enforced by our implicit SDF branch, 3DGS provides a more suitable explicit representation for dynamic tracking in our framework.

\begin{table}[t]
\centering
\caption{\textbf{Impact of Gaussian representation on the D-NeRF benchmark.} We compare our default 3DGS-based model with a variant using 2DGS primitives.
}
\label{tab:3dgs_vs_2dgs}
\resizebox{0.9\linewidth}{!}{
\begin{tabular}{llccccc}
\toprule
Scene & Method
& vIoU $\uparrow$
& CD $\downarrow$
& PSNR $\uparrow$
& SSIM $\uparrow$
& LPIPS $\downarrow$ \\
\midrule

\multirow{2}{*}{Lego}
& Ours-2DGS & 0.2943 & 0.0315 & 25.09 & 0.9374 & 0.0327 \\
& Ours       & \textbf{0.2988} & \textbf{0.0133}
             & \textbf{25.40} & \textbf{0.9439} & \textbf{0.0245} \\
\midrule

\multirow{2}{*}{T-Rex}
& Ours-2DGS & 0.3861 & 0.0286
             & \textbf{36.92} & \textbf{0.9933} & 0.0072 \\
& Ours       & \textbf{0.5783} & \textbf{0.0050}
             & 36.75 & 0.9932 & \textbf{0.0069} \\
\midrule

\multirow{2}{*}{HellWarrior}
& Ours-2DGS & \textbf{0.3057} & 0.0193
             & 31.09 & 0.9821 & 0.0195 \\
& Ours       & 0.3024 & \textbf{0.0124}
             & \textbf{31.14} & \textbf{0.9835} & \textbf{0.0169} \\
\midrule

\multirow{2}{*}{Average}
& Ours-2DGS & 0.3287 & 0.0265 & 31.03 & 0.9709 & 0.0198 \\
& Ours       & \textbf{0.3932} & \textbf{0.0102}
             & \textbf{31.10} & \textbf{0.9735} & \textbf{0.0161} \\
\bottomrule
\end{tabular}
}
\end{table}

\section{Runtime specification}
The Gaussian Splatting branch is trained for 2 hours to capture the initial dynamic topology. Subsequently, the implicit VoxGS-DSDF branch is optimized for another 2 hours to build the continuous SDF field. Finally, the joint mesh refining stage requires approximately 1 hour to converge and output the final temporally consistent meshes.

\section{Limitation and Future Work}
While DySurface effectively extracts high-fidelity meshes from dynamic scenes, there exist limitations regarding extreme topological transformations. The Stage 2 implicit SDF presents a clean surface; minor artifacts arise during Stage 3 mesh refinement. Fitting fine-grained photometric losses during non-rigid dynamics can induce minor vertex noise and normal shifts. Increasing Laplacian regularization ($\mathcal{L}_{lap}$) in Stage 3 suppresses these artifacts; however, this may reduce fine-grained geometric details.

In our framework, we define a specific time step $t$ as the canonical space; however, this single canonical space might not contain sufficient context of the object's surface that undergoes severe fragmentation or topology changes. Extending our approach to effectively model topology-breaking dynamics, such as fracturing or fluid splashing, remains an important direction for future research.

Additionally, while our method excels on 360-degree captures with clear foreground separation, unconstrained in-the-wild videos with severe single-view occlusions receive insufficient photometric supervision, causing localized surface over-smoothing.

\section{Discussion of Related Dynamic Surface Reconstruction Methods}
\label{sec:related_discussion}

\paragraph{Comparison with DGNS.}
DGNS~\cite{sdgns} also combines deformable 3D Gaussian Splatting with a neural implicit surface representation. However, its formulation does not explicitly address the structural discrepancy between the forward deformation of Gaussian primitives and the backward mapping required for volumetric implicit rendering. In contrast, DySurface explicitly couples these two deformation directions through KNN-based RayQuery matching over a dynamically voxelized Gaussian representation, together with the SDF-GS anchoring and cycle-consistency objectives. This design directly associates the dynamic Gaussian geometry with the implicit SDF queries and encourages bidirectional alignment between the explicit and implicit representations. In the qualitative results reported by DGNS, the reconstructed surfaces tend to appear smoother with limited high-frequency geometric details, whereas DySurface preserves finer surface structures.

\paragraph{Comparison with DynaSurfGS.}
DynaSurfGS~\cite{sdynasurfgs} reconstructs dynamic surfaces using planar Gaussian surfels based on a 2DGS [17] representation. Such surface-oriented primitives provide an explicit geometric representation, but their planar parameterization can be restrictive under complex non-rigid deformation. DySurface instead adopts deformable 3D Gaussian primitives, which provide full 3D covariance for modeling dynamic motion, while delegating continuous surface regularization to a multi-resolution hash-grid SDF. The resulting implicit surface is subsequently converted into a canonical mesh, to which the learned deformation field is transferred for temporal articulation. Qualitatively, the meshes reported by DynaSurfGS appear relatively smooth and contain fewer fine-scale surface details, while DySurface better preserves high-frequency geometric structures.

\section{Statistical Significance}
To ensure the reliability and robustness of our quantitative evaluation, we report the statistical significance of the results. For each experimental setting, we independently perform the evaluation 10 times. All values reported in the quantitative tables represent the average (mean) across these 10 runs, accompanied by their corresponding standard deviations ($\pm$).

\begin{table}[htbp]
\centering
\caption{\textbf{Quantitative results of statistical significance.} All reported values represent the mean and standard deviation ($\pm$) computed over 10 independent evaluation runs.}
\label{tab:statistical}
\begin{adjustbox}{scale=0.93}
\begin{tabular}{c|cc|ccc}
\toprule
\:\: Methods  \:\:  & \: vIoU (↑) \: & \: CD (↓) \: & \: PSNR (↑)  & SSIM (↑)  &  \: LPIPS (↓) \: \\ \midrule
GaGS     & $0.2422 {\scriptstyle \pm 0.0031}$ & $0.0146 {\scriptstyle \pm 0.0006}$ & $\textbf{32.24} {\scriptstyle \pm 0.04}$ & $\textbf{0.9753} {\scriptstyle \pm 0.0007}$ & $0.0181 {\scriptstyle \pm 0.0002}$ \\ 
DG-Mesh  & $0.3814 {\scriptstyle \pm 0.0009}$ & $0.0137 {\scriptstyle \pm 0.0002}$ & $25.23 {\scriptstyle \pm 0.04}$ & $0.9186 {\scriptstyle \pm 0.0008}$ & $0.1027 {\scriptstyle \pm 0.0004}$ \\
D-2DGS  & $0.3426 {\scriptstyle \pm 0.0021}$ & $0.0111 {\scriptstyle \pm 0.0004}$ & $25.82 {\scriptstyle \pm 0.05}$ & $0.9374 {\scriptstyle \pm 0.0006}$ & $0.0681 {\scriptstyle \pm 0.0003}$ \\
Ours     & $\textbf{0.3932} {\scriptstyle \pm 0.0005}$ & $\textbf{0.0102} {\scriptstyle \pm 0.0002}$ & $31.09 {\scriptstyle \pm 0.04}$ & $0.9735 {\scriptstyle \pm 0.0016}$ & $\textbf{0.0161} {\scriptstyle \pm 0.0004}$ \\ \bottomrule
\end{tabular}
\end{adjustbox}
\end{table}

\section{Application Example}
\begin{figure*}[htbp]
  \centering
  \includegraphics[width=\textwidth]{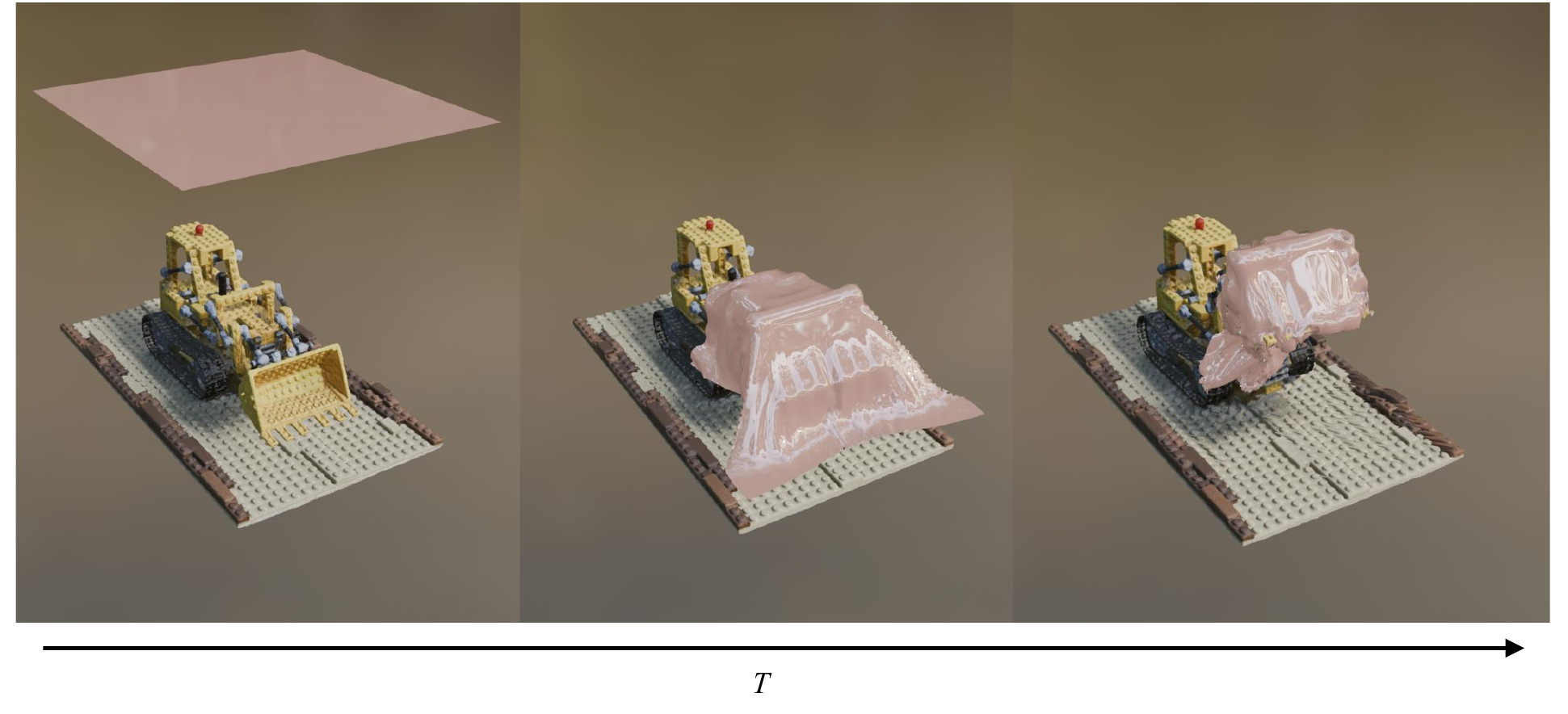}
  \caption{\textbf{Illustration of a DySurface application.} We demonstrate an example downstream application enabled by our reconstructed dynamic surfaces. The reconstructed excavator mesh sequence serves as an animated kinematic collision boundary, while a separately initialized cloth is simulated as a deformable body interacting with it.}
  \label{sup:fig:application}
\end{figure*}

Mesh remains the predominant representation supported by physics simulators and rendering engines. We illustrate a downstream application of DySurface for physics-based simulation in Fig.~\ref{sup:fig:application}. Since DySurface reconstructs temporally consistent dynamic meshes from image observations, the recovered geometry can be directly integrated into standard simulation engines without manual retopology. Specifically, we import the reconstructed mesh sequence into the simulator as an animated kinematic collision boundary, where vertex positions are driven frame-by-frame by our reconstruction. We then drop a secondary deformable object (a cloth in Fig.~\ref{sup:fig:application}) with assigned material parameters (density, stiffness, damping), and run a forward simulation under gravity. The cloth deforms and responds to collisions with the moving reconstructed surface, demonstrating that DySurface is not only effective for high-fidelity reconstruction but also practical for geometry-aware downstream tasks requiring simulation-ready dynamic surfaces.

\newpage
\begin{figure*}[htbp]
  \centering
  \includegraphics[width=\textwidth]{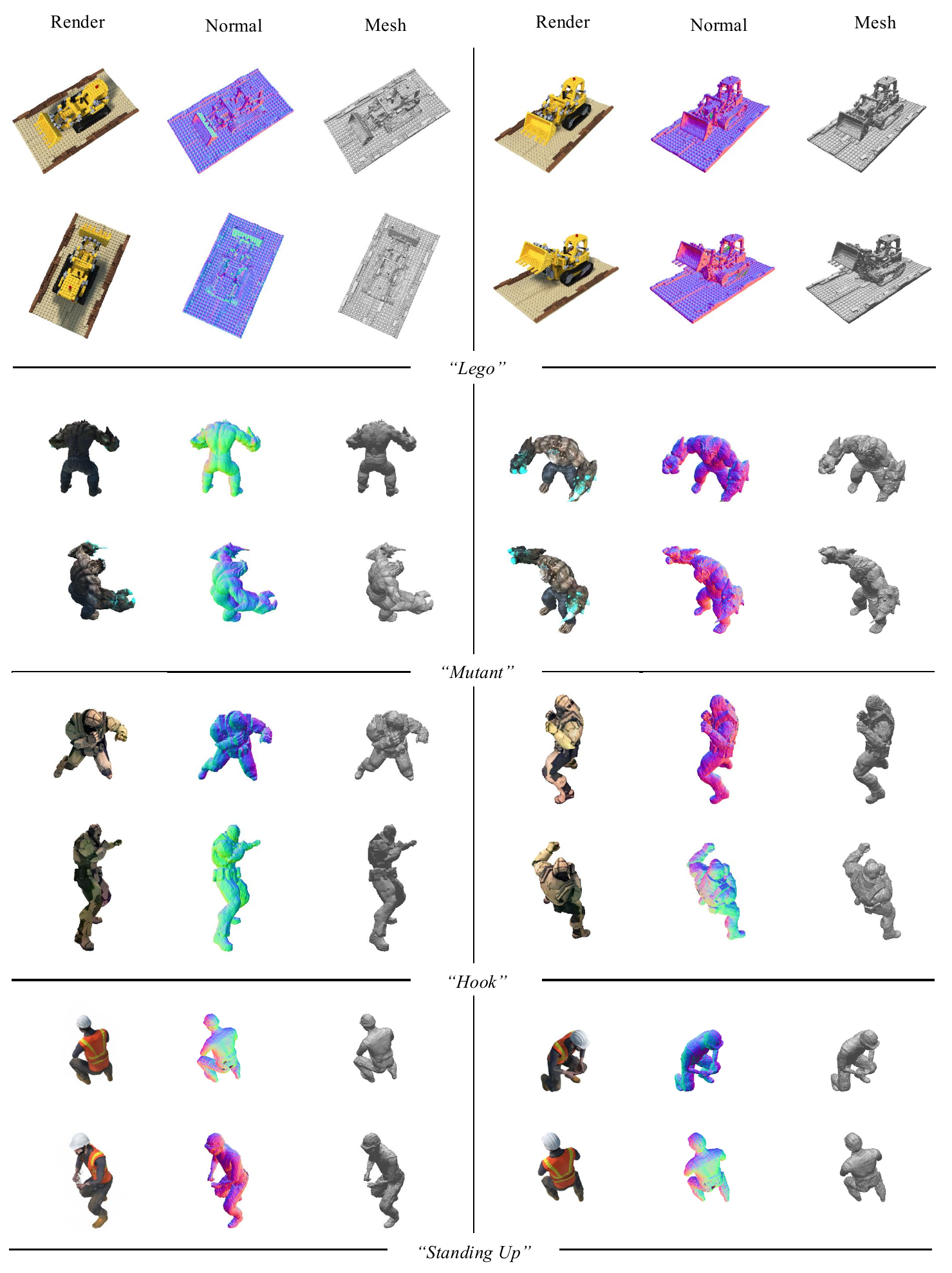}
  \caption{\textbf{Additional qualitative results of diverse D-NeRF scenes.} We show the rendered image, normal maps, and mesh surfaces results of DySurface. Please zoom to check the details.}
  \label{sup:fig:remaining_scenes2}
\end{figure*}

\begin{figure*}[htbp]
  \centering
  \includegraphics[width=\textwidth]{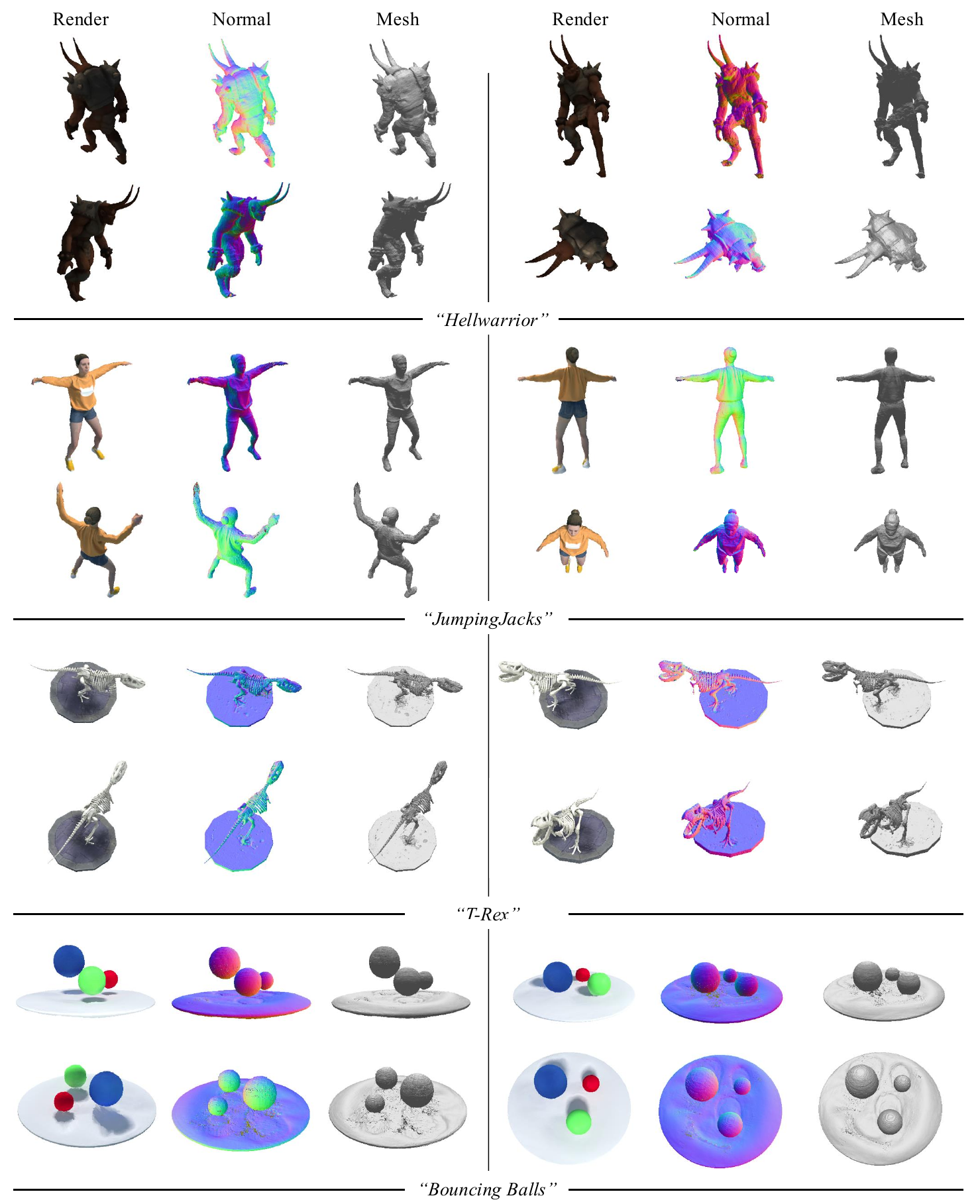}
  \caption{\textbf{Additional qualitative results of diverse D-NeRF scenes.} We show the rendered image, normal maps, and mesh surfaces results of DySurface. Please zoom to check the details.}
  \label{sup:fig:remaining_scenes3}
\end{figure*}

\newpage
{
\small
~
}

%%%%%%%%%%%%%%%%%%%%%%%%%%%%%%%%%%%%%%%%%%%%%%%%%%%%%%%%%%%%
% \newpage
% \input{checklist}

\end{document}